\documentclass[lettersize,journal]{IEEEtran}

\usepackage{amsmath} 
\usepackage{amssymb}  
\usepackage{bm}
\usepackage{graphicx}
\usepackage{xcolor}
\usepackage{booktabs}
\usepackage[labelsep=period]{caption}
\usepackage{subcaption}
\usepackage{algorithmic}
\usepackage[linesnumbered,ruled,vlined]{algorithm2e}
\usepackage{url}
\usepackage{amsthm}
\newtheorem{lemma}{Lemma}
\usepackage{nomencl}
\usepackage{hyperref}

\usepackage{cite}

\ifCLASSINFOpdf

\else

\fi

\begin{document}

\title{Integrating Maneuverable Planning and Adaptive Control for Robot Cart-Pushing under Disturbances}

\author{Zhe Zhang$^\dagger$, Peijia Xie$^\dagger$, Yuhan Pang, Zhirui Sun, Bingyi Xia, Bi-Ke Zhu, Jiankun Wang, \textit{Senior Member}, \textit{IEEE} %
\thanks{$^\dagger$ denotes equal contribution.}
\thanks{Zhe Zhang, Peijia Xie, Yuhan Pang, Zhirui Sun, Bingyi Xia, Bi-Ke Zhu and Jiankun Wang are with SUSTech. Bingyi Xia is also with Peng Cheng Laboratory, Shenzhen, China.}
}

\markboth{Journal of \LaTeX\ Class Files,~Vol.~18, No.~9, September~2020}%
{How to Use the IEEEtran \LaTeX \ Templates}

\maketitle

\begin{abstract}
Precise and flexible cart-pushing is a challenging task for mobile robots. The motion constraints during cart-pushing and the robot's redundancy lead to complex motion planning problems, while variable payloads and disturbances present complicated dynamics. In this work, we propose a novel planning and control framework for flexible whole-body coordination and robust adaptive control. Our motion planning method employs a local coordinate representation and a novel reduced-order model to solve a nonlinear optimization problem, thereby enhancing motion maneuverability by generating feasible and flexible push poses. Furthermore, we present a disturbance rejection control method to resist disturbances and reduce control errors for the complex control problem without requiring an accurate dynamic model. We validate our method through extensive experiments in simulation and real-world settings, demonstrating its superiority over existing approaches. To the best of our knowledge, this is the first work to systematically evaluate the flexibility and robustness of cart-pushing methods in experiments. The supplmentary videos and material are available at \url{https://sites.google.com/view/mpac-pushing/}.
\end{abstract}


\IEEEpeerreviewmaketitle

\section{Introduction}
\label{sec:intro}

\IEEEPARstart{C}{art-pushing} is common in daily life, such as in cargo transportation, shopping assistance and luggage handling at airports. Human operators can fully exploit the redundancy of their two arms to generate flexible push poses, allowing them to maneuver carts through narrow and dynamic spaces while adaptively adjusting their control forces to counteract disturbances such as payload variations and ground resistance. However, this remains challenging for robots. Prior studies \cite{tan2001integrated,tan2001unified,tan2003integrated,xiao2022robotic,xia2023collaborative,xie2025autonomous} have employed mobile robots to
replace humans in cart-pushing task, but their simplified robot designs and control strategies often struggle in real-world environments that are highly dynamic and complex. Subsequent works have explored the use of redundant Degrees of Freedom (DoFs), and dual-arm mobile robots in particular have shown promising control performance due to their human-like structure that suits cart manipulation. However, they still lack flexibility and robustness. A method that fully leverages redundancy for maneuverable planning and maintains robust control against diverse disturbances is still needed. 

\begin{figure}
\centerline{\includegraphics[width=\columnwidth]{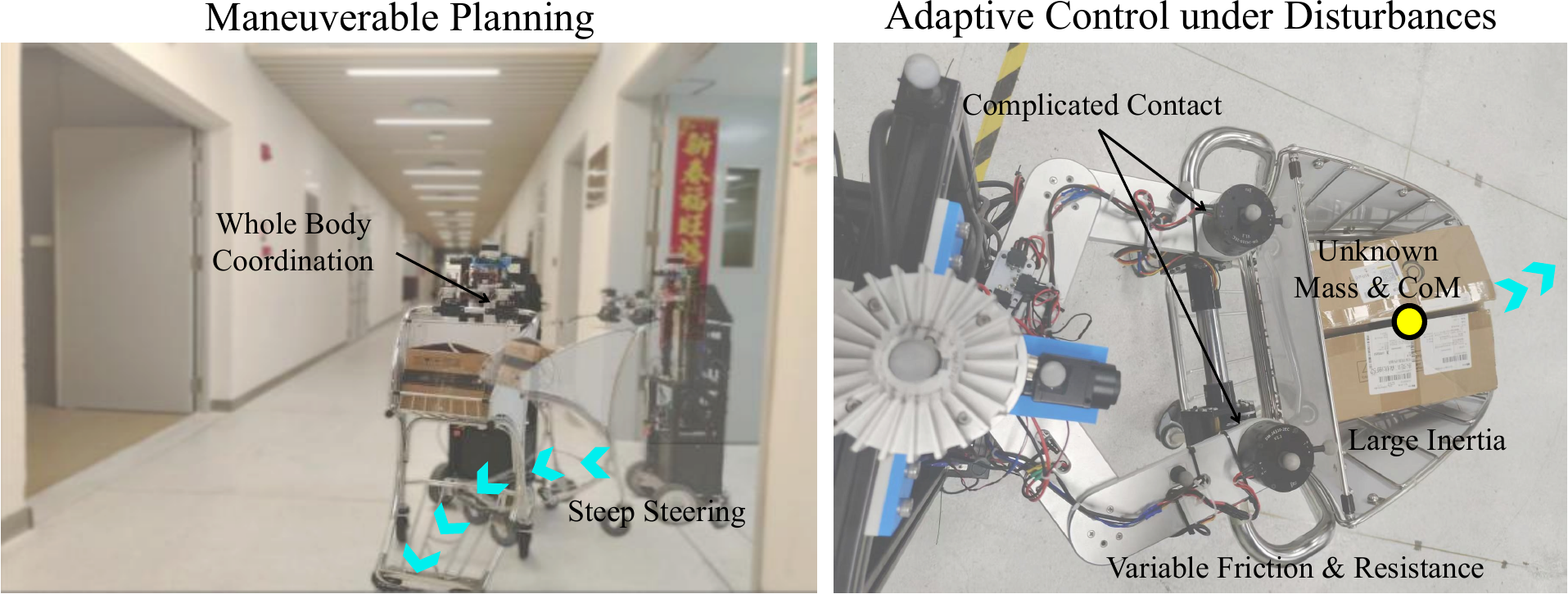}}
\caption{Illustration of maneuverable planning and adaptive
control framework for cart-pushing task. The robot coordinates whole-body motions to maneuver in tighter and more complex environments, while adaptive control enables it to handle various disturbances during cart-pushing.}
\label{fig:illustration}
\end{figure}

The limited local planning of robots is a major factor restricting maneuverability during cart-pushing. Most existing works focus solely on global planning and ignore whole-body coordination. While this may suffice in open environments, the resulting large turning radius makes robot deployment difficult in narrow and dynamic spaces. Early works
\cite{tan2001integrated,tan2001unified,tan2003integrated}  analyzed the nonholonomic model of cart-pushing robot and design a kinodynamic planning method. Our previous works \cite{xiao2022robotic,xia2023collaborative,xie2025autonomous} designed a mobile robot with a single-link manipulator for automated luggage cart retrieval and transportation in airports, but they primarily focused on global planning. Subsequent works employed redundant manipulators such as robotic arms or dual-arm  to enhance maneuverability, but the redundancy makes the planning and control complex. \cite{scholz2011cart, schulze2023trajectory} introduced reduced-order model to minimally parameterize the control inputs of the dual-arm mobile robot, and design corresponding global planning algorithms for cart-pushing but were limited to offline computation. \cite{dai2024wheelchair} further developed an online pushing pose optimizer based on Quadratic Programming (QP), however, it was only capable of tracking predefined velocity trajectories instead of simultaneously performing global path planning and push-pose optimization. These works also did not fully leverage redundancy to generate more flexible push poses. Compared to these works, we firstly build an unconstrained local coordinate representation to minimally represent the redundant dual-arm system with multiple constraints, and propose a whole body motion planning method named Leader-Follower Model Predictive Control (LF-MPC) for simultaneous push-pose optimization and motion planning. LF-MPC can fully leverage redundant DoFs through whole-body motion coordination and generate feasible trajectories online, thereby enhancing cart-pushing maneuverability.

To handle the various complex disturbances during cart pushing, online adaptive control is also essential. However, most existing works either overlook this aspect or rely on offline dynamics identification. \cite{aguilera2023control} and \cite{aguilera2023modeling} proposed a Linear Quadratic Regulator (LQR) controller for cart-pushing, where the inertial parameters were estimated using an Extended Kalman Filter (EKF). \cite{ren2024whole} introduced a controller based on a Fuzzy Logic System (FLS) to compensate for the dynamic uncertainties of carts, but was heavily rely on modeling accuracy and require extensive manual tuning.


Cart-pushing is a complex control problem characterized by numerous disturbances, significant parameter variations and nonlinear actuation dynamics. It is hard to build the accurate dynamic model of the system. As a result, most existing adaptive control strategies \cite{khalil1996robust, nguyen20151, sombolestan2021adaptive, li2023autotrans} are not practical for this task. Other adaptive control strategies \cite{roy2002adaptive, tung2000application} and disturbance observers (DoBs) \cite{chen2000nonlinear, han2009pid, johnson1970further, sira2012robust, zhang2024eso, li2024fixed, wang2024gob} that do not rely on, or depend only weakly on, model accuracy, provide valuable insights. Dual-loop Proportional Derivative (PD) control \cite{roy2002adaptive} and Model Reference Adaptive Control (MRAC) \cite{tung2000application} 
 are efficient adaptive control strategies and are widely used in robotics. Other methods employ DOBs to online estimate disturbances for feedforward control. The extended state observer (ESO) \cite{han2009pid}, unknown input observer (UIO) \cite{johnson1970further}, and generalized proportional–integral observer (GPIO) \cite{sira2012robust} do not rely on accurate dynamic modeling and can achieve
zero estimation error in linear systems. Although these methods performs well in linear systems, experiment results show that they exhibit large steady-state errors and phase lag, and have difficulty to deploy in highly constrained nonlinear cart-pushing system. To address this issue, we develop a nonlinear disturbance observer (NDOB), denoted as the Geometric Observer (GOB), on the proposed local coordinates for feedforward compensation. We further introduce a force-projection strategy that distributes the compensated feedforward forces into the joint torques of the dual-arm manipulators.

Thus, the contributions and novelties can be summarized as follows:

1) We propose a whole-body motion planning algorithm for maneuverable cart-pushing. To deal with the inherent complexity of the constrained redundant system, we present a local coordinate representation of the control inputs, which reduce the problem dimension and eliminate the complex constraints. Then we propose LF-MPC, an online optimization-based method for simultaneous motion planning and push pose optimization. The LF-MPC builds on a reduced-order LF model that effectively exploits system redundancy for efficient local planning.

2) We design a robust adaptive controller for online disturbance compensation without requiring accurate dynamics. The control strategy employs GOB to estimate disturbances in the proposed local coordinate, followed by a force-projection scheme that maps the feedforward control input from the local coordinates to the joint space.


3) We deploy the planning and control framework on our designed dual-arm mobile robot and validate its performance through extensive simulations and real-world experiments.

To the best of the authors’ knowledge, this is the first work to systematically analyze and experimentally validate the flexibility and robustness requirements in cart-pushing tasks.

The rest of this article is organized as follows. Section \ref{sec:problem} discusses the task requirements and challenges of cart-pushing. Sections \ref{sec:hardware} introduce the hardware platform. Sections \ref{sec:planning_module} and \ref{force_sec} outline the planning and control algorithm, respectively. In Section \ref{exp_sec}, we present experimental results, and finally, Section \ref{conclusion} concludes the article.

\begin{figure}[!t]
\centerline{\includegraphics[width=\columnwidth]{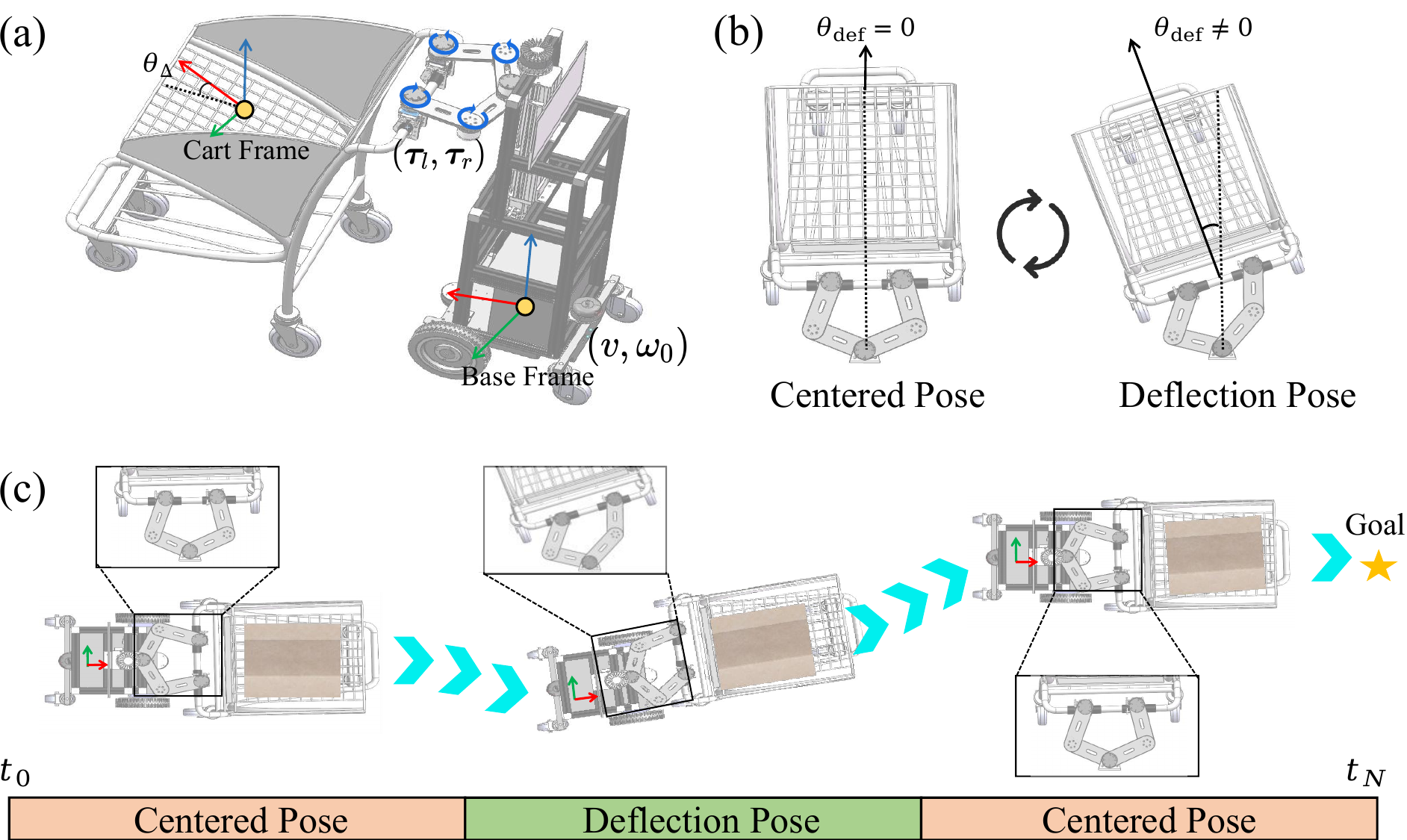}}
\caption{(a) Control inputs of arms and base. (b) Two distinct
push poses. (c) Illustration of whole body coordination requirement. The planner should flexibly utilize Deflection Poses to enhance maneuverability and ultimately return to a stable Centered Pose during transportation.}
\label{fig:task_requirements}
\end{figure}

\section{Problem Formulation}
\label{sec:problem}

\subsection{Notations}
In the rest of the paper, for convenient representation we define: $\theta$ denotes the yaw angle, and $\omega$ denotes its angular velocity. $\boldsymbol{s}_{\star}=[x_{\star},y_{\star},\theta_{\star}]^\top$ represents the pose of $\star$, $\boldsymbol{p}_{\star}= [x_{\star},y_{\star}]^\top$ is the position, and $p_{\star}$ is the exact point. $\boldsymbol{R}, \boldsymbol{T}$ are the rotation and transformation matrix. $\boldsymbol{\mu}_{\star}=[v_{\star}, \omega_{\star}]^{\top}$ is the velocity vector. $\boldsymbol{s}_d, \boldsymbol{s}_0, \boldsymbol{s}_{\text{c}}, \boldsymbol{s}_l, \boldsymbol{s}_r$ denote the desired pose, the pose of the mobile base, the geometric center of the cart, the left and right gripper, respectively. $^{\text{W}}\boldsymbol{s}, ^{\text{B}}\boldsymbol{s}, ^{\text{C}}\boldsymbol{s}$ denote the pose represented in the world frame, mobile base frame and cart frame. The subscripts of the other variables are defined in the same manner. 


\subsection{Description of the Task}

We focus on the local planning and control of cart-pushing tasks. Given the global path (e.g., sparse waypoints or continuous reference trajectories) $\boldsymbol{S}_d=[\boldsymbol{s}_d^0,\dots, \boldsymbol{s}_d^k]^{\top}$  as high-level intentions, where $\boldsymbol{s}_d=[x_d, y_d, \theta_d]^{\top}$ is the desired cart pose in the world frame, the objective is to compute the optimal control inputs of arms and mobile base to push and steer the cart to accurately track $\boldsymbol{S}_d$, while satisfying multiple constraints. The overall kinodynamic planning problem for the redundant robot is constructed as follows:
\begin{equation}
    \begin{aligned}
    \min_{\boldsymbol{u}_k} \quad & \sum_{k=0}^{N-1} (\tilde{\boldsymbol{x}}_k^{\top}\boldsymbol{Q}_k\tilde{\boldsymbol{x}}_k+\boldsymbol{u}_k^{\top}\boldsymbol{R}_k\boldsymbol{u}_k)+\tilde{\boldsymbol{x}}_N^{\top}\boldsymbol{Q}_N\tilde{\boldsymbol{x}}_N \\
    \text{s.t.} \quad & \boldsymbol{x}_{k+1}=\boldsymbol{x}_k+f(\boldsymbol{x}_k, \boldsymbol{u}_k)\Delta t \\
    & \boldsymbol{x}, \boldsymbol{u} \in \mathcal{C} 
    \end{aligned}\label{optimization}
\end{equation}
where $\tilde{\boldsymbol{x}}_k=\boldsymbol{x}_d-\boldsymbol{x}_k$, $\boldsymbol{x}_d = \boldsymbol{s}_d$, $\boldsymbol{x}_k=\boldsymbol{s}_c$ and $\boldsymbol{s}_c$ is the current cart pose. $\boldsymbol{u}_k$ is the system control inputs. $N$ is the planning horizon, $\boldsymbol{Q}_k, \boldsymbol{Q}_N$ and $\boldsymbol{R}$ are positive definite matrices, $f$ is the nonlinear state transition function and $\mathcal{C}$ is a convex set, e.g., input bounds and constraints. Some works define $\boldsymbol{u}_k$ as joint torques and solve (\ref{optimization}) to perform the entire planning and control, but uncertainties in the dynamics can degrade performance. We therefore consider a hierarchical framework in which $\boldsymbol{u}_k$ represents the desired displacement, while the joint torques are generated by an adaptive controller. The problem can be solved by a nonlinear optimizers but is highly complex due to redundancy and complicated constraints. Our goal is to develop a general framework for maneuverable motion generation and robust control of dual-arm mobile robot in cart-pushing task.

\subsection{Task Requirements}

Building on the task description, we further specify a set of requirements for cart-pushing tasks to achieve improved control performance and discuss the associated challenges.

\subsubsection{Complex Constraints Satisfaction}\label{constraintformulation}

The task involves multiple constraints. For the dual-arm manipulator, the grippers continuously hold the cart handle, forming a polygonal structure that imposes geometric equality constraints. In addition, physical bounds of the arms introduce inequality constraints. For the differential drive mobile base, the imposed nonholonomic constraints are expressed as:
\begin{equation}
 \dot{x}=v\cos \theta, \dot{y}=v\sin \theta, \dot{\theta}=\omega
\label{nonholonomic}
\end{equation}
where $v$ and $\omega$ are the longitudinal and angular velocity,  and $\theta$ is the orientation. The above constraints make (\ref{optimization}) complex and difficult to solve in real time.

\subsubsection{Whole-Body Coordination}\label{requirement}

The robot’s redundancy leads to multiple feasible solutions for (\ref{optimization}), yet fully exploiting this redundancy is crucial for maneuverability. In particular, the robot must use the redundant DoFs of its dual arms to reduce the turning radius for navigating narrow spaces (Deflection Pose), while maintaining a well-balanced Centered Pose at other times to enhance control robustness. Effective whole-body coordination requires flexible redistribution of control effort between the arms and the base, alternating between different push poses to reduce tracking error.
Directly solving (\ref{optimization}) cannot achieve this, as it is prone to local minima. Although some works restrict the DoFs to simplify the problem, we seek an approach that solves (\ref{optimization}) efficiently without sacrificing redundancy.


\subsubsection{Control under Complicated Dynamics}

The system is highly nonlinear and subject to numerous constraints, and the disturbances are difficult to model, making the design of an effective control strategy challenging. The dynamics of carts are influenced by various factors, including payload mass variations, shifts in the CoM due to payload movement, friction and wheel resistance, and changes in the robot’s contact state (e.g., grasping locations). Additionally, wheel resistance is highly uncertain, and friction can vary significantly depending on the ground surface and payload conditions. These effects are difficult to accurately observe and model, resulting in high uncertainty in dynamics. Our goal is to achieve robust adaptive control for online compensation, without relying on accurate modeling of the system dynamics.

\section{Hardware Platform}\label{sec:hardware}

\begin{figure}[!t]
\centerline{\includegraphics[width=0.9\columnwidth]{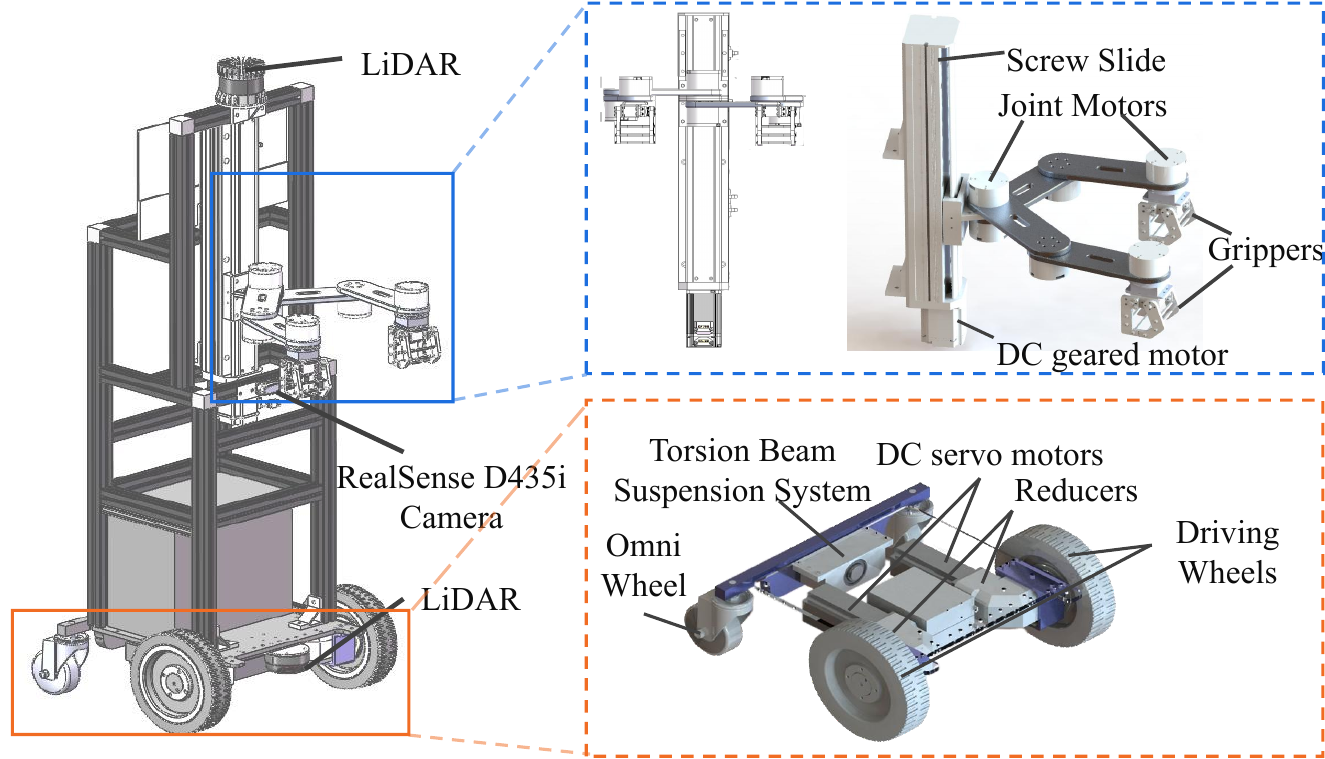}}
\caption{Hardware platform overview.}
\label{hardware}\vspace{-2mm}
\end{figure}

The hardware platform is a mobile robot similar to that used in our previous works \cite{xiao2022robotic, xia2023collaborative, xie2025autonomous}. It consists of a perception module equipped with a 32-line Ouster OS0 LiDAR and Realsense cameras, and an actuation module comprising a CAN box and a differential-drive mobile base. Unlike the previous setup, the manipulators are two SCARA robotic arms driven by six DC motors, as shown in Fig.~\ref{hardware}. The robot arm actuators operate under the MIT protocol, enabling hybrid force–position control. The CAN bus runs at 120 Hz. The dual arms can move vertically along a linear stage to accommodate carts of different heights. 

\vspace{2mm}
\section{Whole-Body Motion Planning}\label{sec:planning_module}

In this section, we present an efficient motion planner that fully exploits the system’s redundancy (as shown in the middle of Fig. \ref{fig:framework}). Given a global path as a reference trajectory (generated by Hybrid A*), our planner enables online whole-body motion planning to generate flexible push pose and track the trajectory. We first project the high-dimensional, constrained control inputs of the dual arms onto a low-dimensional, unconstrained local coordinate. We then design LF-MPC to perform push-pose optimization and local planning.


\subsection{Local Coordinate Representation}\label{sec:local_coordinate}

Although the polygon representation in Section \ref{constraintformulation} simplifies the constraints, planning with a deformable polygon remains challenging. Additionally, for robotic arms with variable configurations and DoFs, the polygon’s shape and number of edges must be adapted accordingly, further complicating tuning. Note that the intrinsic dimension of a constrained redundant system may be lower than that of the original formulation. Therefore, we explore a more efficient local coordinate representation that minimally parameterizes the redundant dual-arm system while eliminating the constraints.

\begin{figure}[!t]
\centerline{\includegraphics[width=0.9\columnwidth]{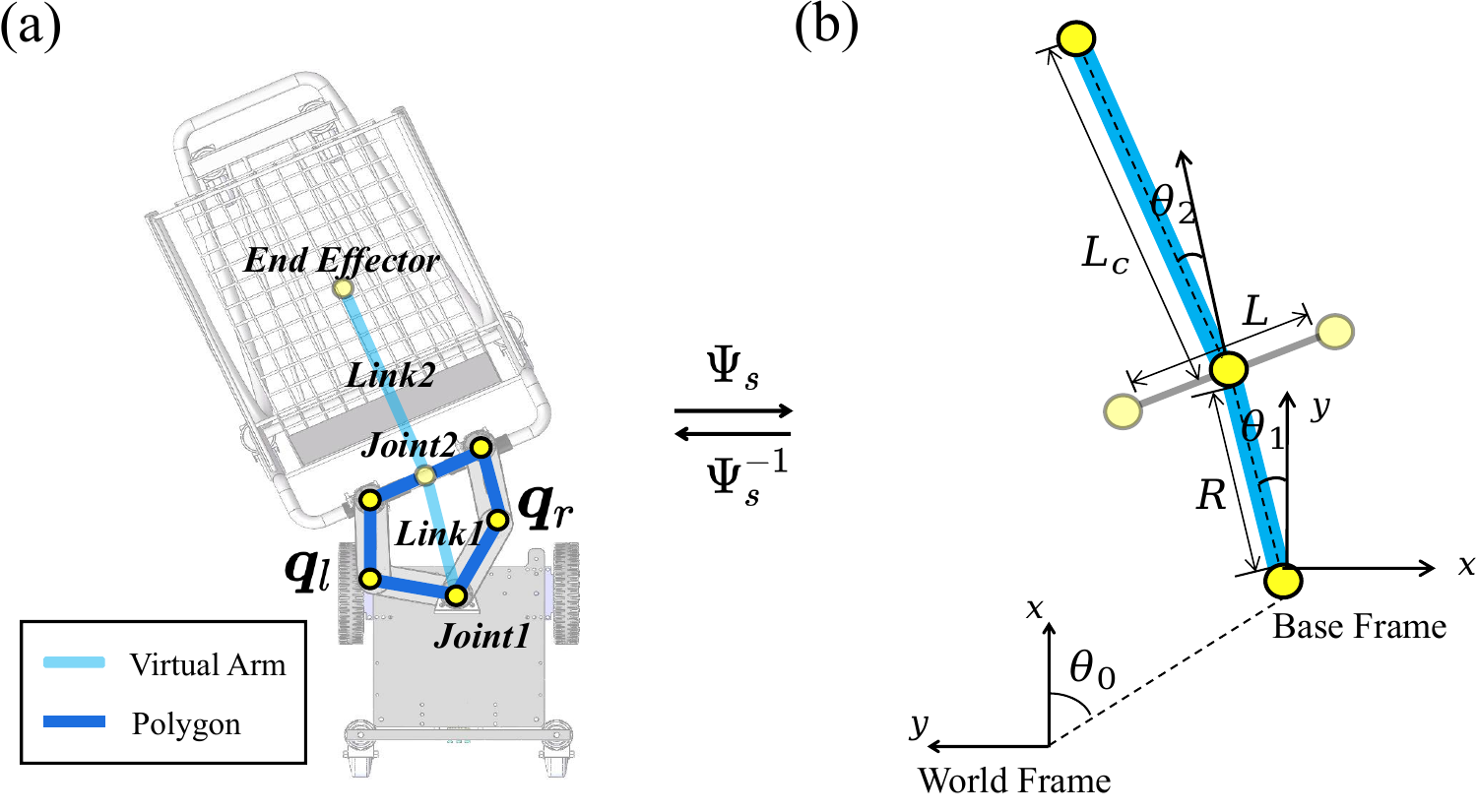}}
\caption{Local coordinate representation of arm states. (a) The original system. (b) The virtual 2-link robotic arm model.}
\label{local_coordinates}
\end{figure}

Assuming the grippers achieve sufficiently symmetrical grasping (a reasonable assumption since most carts feature symmetrical handles and visual methods enable accurate symmetrical grasping), we can further simplify the polygonal constraint into a two-link virtual arm model, as illustrated in Fig. \ref{local_coordinates}. In this model, \textit{Joint1} is located at the arms' base link, \textit{Joint2} at the center between the grippers, and the \textit{End-Effector} (\textit{EE}) at the geometric center of the cart. Here, $\theta_1$ and $\theta_2$ denote the orientation angles of \textit{Joint1} and \textit{Joint2} respectively, $R$ is the length of \textit{Link1}, $L_c$ is the length of \textit{Link2} and $L$ is the distance between grippers. The grippers are treated as attachments to \textit{Joint2}. With this construction, the dual-arm–cart configuration can be fully parameterized by a single virtual arm. We use it as the local coordinate representation and it inherently satisfies the polygon constraints. We therefore define the control inputs in the local coordinate system as $\boldsymbol{\psi}=[\theta_1,\theta_2,R,L]^\top\in\mathbb{R}^4$. It is easy to derive the mapping function $\Psi_s: \boldsymbol{\psi} \mapsto \boldsymbol{q}$ and $\Psi_s^{-1}: \boldsymbol{q} \mapsto \boldsymbol{\psi}$ as:
\begin{equation}
    \Psi_s: \boldsymbol{\psi}=g(\boldsymbol{s})=g(f_{\mathrm{FK}}(\boldsymbol{q})), 
\Psi_s^{-1}: \boldsymbol{q}=f_{\mathrm{IK}}(\boldsymbol{s})=f_{\mathrm{IK}}(g^{-1}(\boldsymbol{\psi})),
\end{equation}
where $f_{\text{IK}}(\cdot)$ and $f_{\text{FK}}(\cdot)$ are inverse kinematic and forward kinematic functions, and $g$ denoted as the mapping $g: \boldsymbol{s} \mapsto \boldsymbol{\psi}$. For 3-DoF arms, $f_{\text{FK}}$ and $f_{\text{IK}}$ are unique. For redundant arms, uniqueness can also be achieved by optimizing additional objectives (e.g., minimizing energy). We next discuss whether the mapping $g$ is unique. We define a base frame with \textit{joint1} as the origin. Once the mobile base’s pose in the world frame is specified, the projection from the world frame to the base frame becomes straightforward. The SE(2) pose in the base frame is represented as:
\begin{equation}
^{\text{B}}\boldsymbol{T}_\star=
\begin{bmatrix}
\boldsymbol{R}(\theta_{\star}) & \boldsymbol{t}_{\star} \\
\boldsymbol{0} & \boldsymbol{1}
\end{bmatrix}.
\end{equation}
in which $^{\text{B}}\star$ denotes $\star$ in the mobile base frame. Denoting the SE(2) pose of \textit{Joint2}, the left and right gripper as $^{\text{B}}\boldsymbol{T}_c, ^{\text{B}}\boldsymbol{T}_l$ and $^{\text{B}}\boldsymbol{T}_r$. Then we have:
\begin{equation}
\boldsymbol{R}(\theta_{c})=\boldsymbol{R}(\theta_{l,r})=
\begin{bmatrix}
\cos\theta_{\Delta} & -\sin\theta_{\Delta} \\
\sin\theta_{\Delta} &  \cos\theta_{\Delta}
\end{bmatrix},
\end{equation}
in which $\theta_{\Delta}=\theta_1+\theta_2$, and
\begin{equation}
    \mathbf{t}_c = 
\begin{bmatrix}
R\sin\theta_1 + L_c\sin\theta_{\Delta} \\
R\cos\theta_1 + L_c\cos\theta_{\Delta}
\end{bmatrix},
\mathbf{t}_{l,r}=
\begin{bmatrix}
R\sin\theta_1 \mp 0.5L\sin\theta_{\Delta} \\
R\cos\theta_1 \pm  0.5L\cos\theta_{\Delta}
\end{bmatrix}.
\end{equation}
So $g: \boldsymbol{s}_{l,r} \mapsto 
 {^{\text{B}}\boldsymbol{T}_{l,r}} \mapsto \boldsymbol{\psi}$ is unique and its inverse $g^{-1}$ is evidently unique as well.

With the local coordinate mapping, the states in motion planning problem becomes $\boldsymbol{x} = [\boldsymbol{s}_c, \boldsymbol{\psi}]^{\top}$ and $\boldsymbol{u}$ becomes $[\boldsymbol{\mu}_0, \dot{\boldsymbol{\psi}}]^{\top}$ with $\boldsymbol{\mu}_0=[v_0, \omega_0]^{\top}$. We ignore $R$ and $L$ since they have little impact on cart transportation, and the $\boldsymbol{x}$ and $\boldsymbol{u}$ in (\ref{optimization}) becomes $
    \boldsymbol{x}=[x_c, y_c, \theta_c, \theta_1,\theta_2]^{\top}$ and $ \boldsymbol{u}=[v_0,\omega_0, \omega_1, \omega_2]^{\top}$, with $\theta_c = \theta_0+\theta_1+\theta_2$. The local coordinate representation naturally satisfy the polygon constraints and minimally parameterize the control inputs, thereby simplify the problem but note that (\ref{optimization}) is still hard to solve. Directly solving the full-order problem is hard due to the problem described in Section \ref{requirement}. On the one hand, in narrow environments, reducing the turning radius requires relying more on arm motions (i.e., $\omega_1$ and $\omega_2$). On the other hand, stable transportation requires avoiding excessive use of arm motions. Relying solely on weight tuning makes it difficult to manage the trade-off between the requirements of flexibility and stability, and can easily lead to local optima. We now seek for a proper reduced-order model $f$ to improve the tractability of the optimization problem.


\subsection{Stable Planning\textemdash{}Truck and Trailer MPC}\label{stablempc}

\begin{figure}[!t]
\centerline{\includegraphics[width=\columnwidth]{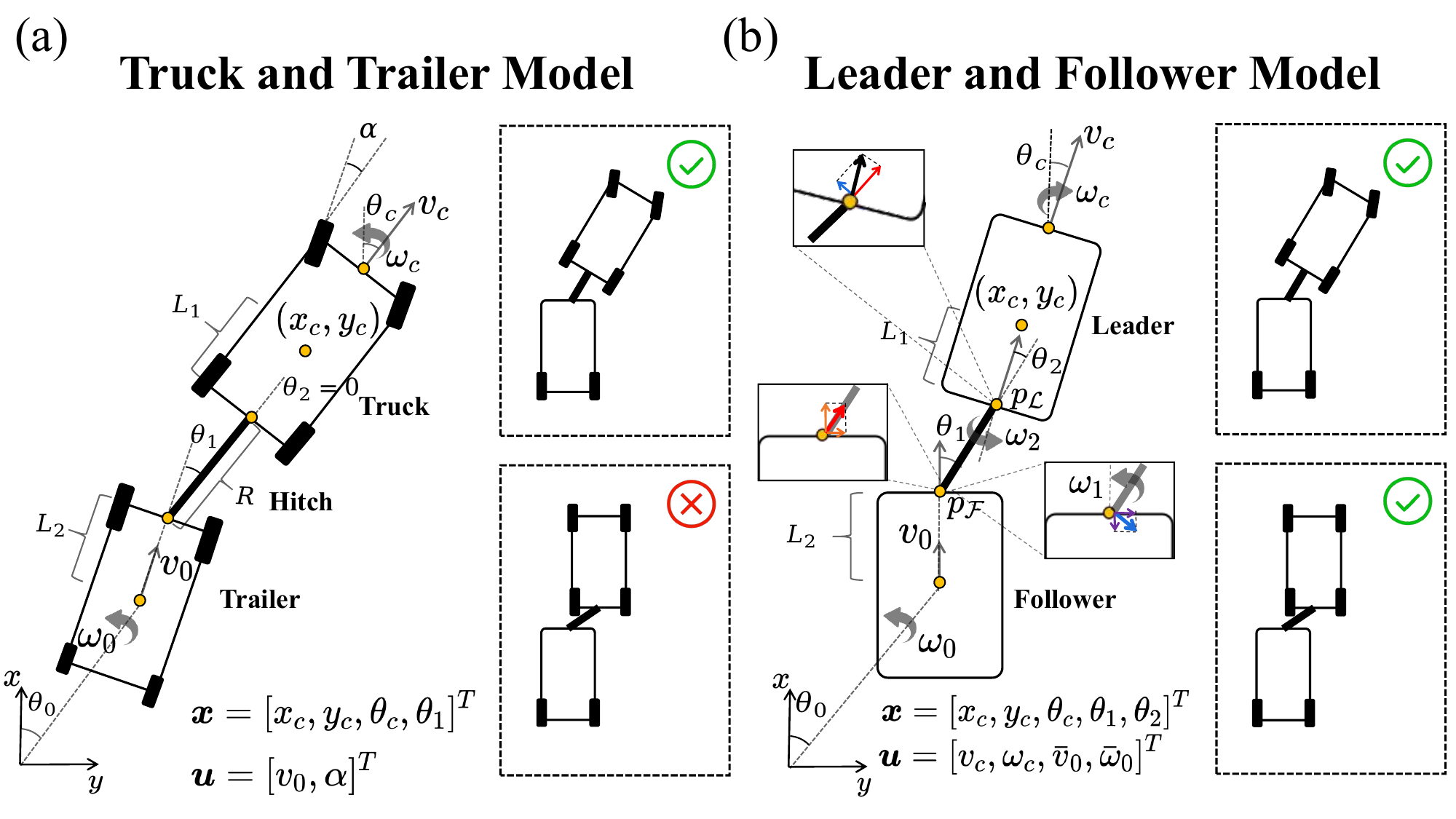}}
\caption{Illustration of the reduced-order models. (a) Global view of the TT model. TT model cannot achieve poses where the truck shifts laterally relative to the trailer. (b) Global view of the LF model. LF model can fully leverage its maneuverability to execute complex motions while fulfilling the task requirements.}
\label{LnF}
\end{figure}

Before presenting our model, we first introduce a simple yet effective model. We observe that the Truck–Trailer (TT) system naturally satisfies these requirements: the hitch introduces an additional DoF between the truck and the trailer, enabling a smaller turning radius. Moreover, once the turn is completed, the hitch automatically pulls the trailer back to a stable centered pose. Inspired by the TT model, we model the cart and mobile base as a virtual truck and trailer, with the manipulator acting as the hitch. This formulation inherently aligns with the motion local coordinates representation in Section \ref{sec:local_coordinate}. In TT model,  the state and control inputs becomes $\boldsymbol{x}=[x_c, y_c, \theta_0, \theta_1]^{\top}, \boldsymbol{u}=[v_0,\alpha]^{\top}$,
in which $\alpha$ is the truck's steering angle and $\theta_2 \equiv 0$. We can derive that:
\begin{equation}
    f_{\text{TT}}: \dot{\boldsymbol{x}} =\begin{bmatrix}
v_c\cos\theta_1(1+\frac{R}{L_1}\tan\alpha \tan \theta_1)\cos \theta_0
 \\
v_c\cos \theta_1 (1+\frac{R}{L_1}\tan\alpha\tan\theta_1)\sin\theta_0
 \\
v_c(\frac{\sin\theta_1}{L_2} -\frac{R}{L_1L_2}\cos \theta_1\tan \alpha)
 \\
v_c(\frac{\tan\alpha}{L_1}-\frac{\sin\theta_1}{L_2}+\frac{R}{L_1L_2}\cos\theta_1\tan\alpha)
\end{bmatrix}\label{ttmodel}
\end{equation}
where $L_1$ is the half of the virtual truck length and $L_2$ is the virtual trailer length as shown in Fig. \ref{LnF} (a), and
\begin{equation}
v_0=v_c\cos\theta_1(1+\frac{R}{L_1}\tan\alpha \tan \theta_1)
\end{equation}
We denote (\ref{ttmodel}) as $f_{\text{TT}}$ and refer to substituting $f_{\text{TT}}$ into (\ref{optimization}) as TT-MPC. In TT-MPC, planning of $\alpha$ drives the coordinated motion of the arms and the mobile base to achieve system steering, and the configuration automatically returns to the centered pose once the steering ends (\textit{proof} see \href{https://drive.google.com/file/d/1tkd3y1829YD6_nnqEF2RPGybxyoX_pTm/view}{supplementary material}).

TT-MPC is sufficiently flexible for most transportation tasks but does not use the control variable $\theta_2$. Consequently, it cannot achieve the pose shown in the lower right corner of Fig. \ref{LnF} (a), where the truck shifts laterally relative to the trailer, limiting the cart’s inherent maneuverability.

\subsection{Flexible Planning\textemdash{}Leader-Follower MPC}\label{lf-mpc}

Based on the TT model, we present an improved model called the Leader-Follower (LF) model. The LF model introduces an additional rotational DoF (i.e., $\theta_2$) at the connection between the truck and the hitch compared to the TT model, and consider the cart and the robot as the leader and the follower. This enables the system to achieve more complex pose, as shown in Fig. \ref{LnF} (b). The design of the LF model is based on unicycle models linearization method in \cite{zhao2019bearing}: 
\begin{lemma}\label{unicycle_linear}
    Consider a point shifted from the center point $\boldsymbol{p}_r=\boldsymbol{p}+r\boldsymbol{h}$, where $r$ is the distance and $\boldsymbol{h}=[\cos\theta,\sin\theta]^{\top}$, we can derive that:
\begin{equation}
    \dot{\boldsymbol{p}}_r=\begin{bmatrix}\dot{x}_r\\\dot{y}_r\end{bmatrix}=\underbrace{\begin{bmatrix}
 \cos\theta & -r\sin\theta\\
 \sin\theta & r\cos\theta
\end{bmatrix}}_{\boldsymbol{\Lambda}}\underbrace{\begin{bmatrix}v\\\omega\end{bmatrix}}_{\boldsymbol{\mu}} \label{linearize_unicycle}
\end{equation}

In this expression, the offset point removes the nonholonomic constraints and becomes a single integrator capable of lateral sliding motion. Furthermore, the mapping $[\dot{x}_r, \dot{y}_r]^{\top}\mapsto [v,\omega]^{\top}$ is also unique:
\begin{equation}
    \boldsymbol{\mu} =\boldsymbol{\Lambda} ^{-1}\dot{\boldsymbol{p}}_r=\begin{bmatrix}
 \cos\theta & \sin\theta\\
 \frac{-\sin\theta}{r} & \frac{\cos\theta}{r}
\end{bmatrix}\begin{bmatrix}\dot{x}_r\\\dot{y}_r\end{bmatrix}\label{linearize1}
\end{equation}

\end{lemma} 

In the LF model, we consider that the center points of the leader and the follower are subject to nonholonomic constraints. By applying (\ref{linearize_unicycle}) and (\ref{linearize1}), the follower's front point $p_\mathcal{F}$ and the leader's back point $p_{\mathcal{L}}$ becomes unconstrained offset points and satisfy $\dot{\boldsymbol{p}}_{\mathcal{L}}=\boldsymbol{\Lambda}_\mathcal{L} \boldsymbol{\mu}_c$ and $\dot{\boldsymbol{p}}_{\mathcal{F}}=\boldsymbol{\Lambda}_\mathcal{F} \boldsymbol{\mu}_0$. Additionally, given the motion $\boldsymbol{\mu}_c=[v_c, \omega_c]^{\top}$ of the cart, we can derive  $\dot{\boldsymbol{p}}_{\mathcal{F}}$ from $\dot{\boldsymbol{p}}_{\mathcal{L}}$, and subsequently obtain the base motions $\boldsymbol{\mu}_0 =[v_0, \omega_0]^{\top}$ and the arm motions $(\omega_1, \omega_2)$ based on the rigid-body kinematic propagation. The transition function $f_{\text{LF}}$ is formulated as:



\begin{equation}
    f_{\text{LF}}: \begin{bmatrix}\dot{x}_c\\\dot{y}_c\\\dot{\theta}_c\\\dot{\theta}_1\\\dot{\theta}_2\end{bmatrix}
=\begin{bmatrix}
v_c\cos\theta_c
 \\
v_c\sin\theta_c
 \\
 \omega_c
 \\
\omega_c-\omega_0-\omega_2
 \\
\omega_2
\end{bmatrix}
\end{equation}\label{LF-model}
where
\begin{equation}
\begin{cases}
v_0=(\boldsymbol{w}^{+}_{\mathcal{L}})^{\top}\boldsymbol{\mu}_c \cos \theta_1 \\
\omega_0=\frac{2}{L_2}(\boldsymbol{w}^{+}_{\mathcal{L}})^{\top}\boldsymbol{\mu}_c\sin \theta_1 \\
    \omega_2=-\frac{1}{R}(\boldsymbol{w}^{-}_{\mathcal{L}})^{\top}\boldsymbol{\mu}_c  
\end{cases}\label{lfmodel}
\end{equation}
with  $\boldsymbol{w}^{+}_{\mathcal{L}}=[\sin \theta_2, L_1 \cos \theta_2]^{\top}$, $\boldsymbol{w}^{-}_{\mathcal{L}}=[\cos \theta_2, -L_1 \sin \theta_2]^{\top}$. The derivation of (\ref{lfmodel}) is presented in \href{https://drive.google.com/file/d/1tkd3y1829YD6_nnqEF2RPGybxyoX_pTm/view}{supplementary material}. Substituting $f_{\text{LF}}$ into (\ref{optimization}) we obtain LF-MPC. To prevent the motions $\boldsymbol{\mu}_0$ from exceeding the robot's hardware limitations, we further introduce intermediate variables $(\bar{v}_0, \bar{\omega}_0)$ into $\boldsymbol{u}$ and impose inequality constraints on $\boldsymbol{u}$. 

To enable safe and practical cart pushing, collision avoidance is essential. We incorporate discrete-time control barrier function (CBF) for constraints to achieve online collision avoidance. For the elongated-shaped cart and mobile base, to enhance the maneuverability of the cart-pushing system and enable obstacle avoidance in narrow spaces, a suitable collision shape is required for constructing the CBF. Specifically, we use a 2D capsule composed of 2 semicircles of radius $r_{\text{cap}}$ and a rectangle of length $2L_{\text{cap}}$ and width $2r_{\text{cap}}$. Then the CBF $h(\boldsymbol{x})$ is defined as the signed distance function (SDF) of the capsule:
\begin{equation}
    h(\boldsymbol{p}_{\text{obs}}(\boldsymbol{x}_k))=
\sqrt{
\left[x - \mathrm{clip}(x,-L_{\text{cap}},L_{\text{cap}})\right]^2 + y^2
}-r_{\text{cap}}-d_{\text{safe}}, \label{sdf_capsule}
\end{equation}in which $d_{\text{safe}}$ is the safe distance threshold. Note that $\boldsymbol{p}_{\text{obs}}(\boldsymbol{x}_k)=\boldsymbol{T}(\boldsymbol{x}_k){^{\text{W}}\boldsymbol{p}_{\text{obs}}}$ is the obstacle point position translated from world frame to the capsule frame. $L_{\text{cap}}$ and $r_{\text{cap}}$ are the half-length and radius of the capsule. For simplicity, we write CBF as $h(\boldsymbol{x}_k)$. In practice, we model the cart and the mobile base as 2 capsules (denoted as $\text{capsule}_1$and $\text{capsule}_2$), and the capsule centers locate at the center of mobile base $\boldsymbol{p}_0$ and cart $\boldsymbol{p}_{\text{c}}$, respectively. We project the $N_{\text{obs}}$ points closest to the two capsules into the base frame B and the cart frame 
C, respectively, compute the corresponding CBF values using (\ref{sdf_capsule}), and incorporate them into the optimization problem. Defining $\boldsymbol{x}_d = [x_d,y_d,\theta_d,0,0]^{\top}$, $\boldsymbol{x} = [x_c,y_c,\theta_c,\theta_1,\theta_2]^{\top}$ and $\boldsymbol{u}=[v_c, \omega_c,\bar{v}_0,\varpi_0]^{\top}$, the final optimization problem can be formulated as:
\begin{equation}
    \begin{aligned}
    \min_{\boldsymbol{u}_k \in [\boldsymbol{u}_{\text{min}}, \boldsymbol{u}_{\text{max}}]} \quad & \sum_{k=0}^{N-1} (\tilde{\boldsymbol{x}}_k^{\top}\boldsymbol{Q}\tilde{\boldsymbol{x}}_k+\boldsymbol{u}_k^{\top}\boldsymbol{R}\boldsymbol{u}_k)+\tilde{\boldsymbol{x}}_N^{\top}\boldsymbol{Q}_N\tilde{\boldsymbol{x}}_N \\
    \text{s.t.} \quad & \boldsymbol{x}_{k+1}=\boldsymbol{x}_k + f_\text{LF}(\boldsymbol{x}_k, \boldsymbol{u}_k)\Delta t, \\
    & \varpi_0-\frac{1}{r_\mathcal{F}}  (\boldsymbol{w}^{-}_{\mathcal{L}})^{\top}\boldsymbol{\mu}_c \sin \theta_1=0, \\
    & \bar{v}_0-\frac{1}{r_\mathcal{F}} (\boldsymbol{w}_\mathcal{L}^{+})^{\top}\boldsymbol{\mu}_c  \cos\theta_1=0, \\
    & h(\boldsymbol{x}_{k+1})- h(\boldsymbol{x}_k)\ge -\gamma h(\boldsymbol{x}_{k}).
    \end{aligned}\label{optLF}
\end{equation}
in which $\gamma=0.9$ is the decay rate. The LF model allows the robot to fully utilize the arms’ DoFs to generate flexible push poses, thereby improving maneuverability, while still satisfying the task requirements as in the TT model.

\begin{lemma}
    The LF model meets the task requirements in Section \ref{requirement} and will ultimately converge to the centered pose during motion, i.e., $\theta_1 \to 0$ and $\theta_2 \to 0$ as $v_c>0$ and $\omega_c \approx 0$ (\textit{Proof} see \href{https://drive.google.com/file/d/1tkd3y1829YD6_nnqEF2RPGybxyoX_pTm/view}{supplementary material}).
\end{lemma}

\section{Robust Adaptive Controller} \label{force_sec}

\begin{figure*}
   \centering  \includegraphics[width=0.9\linewidth]{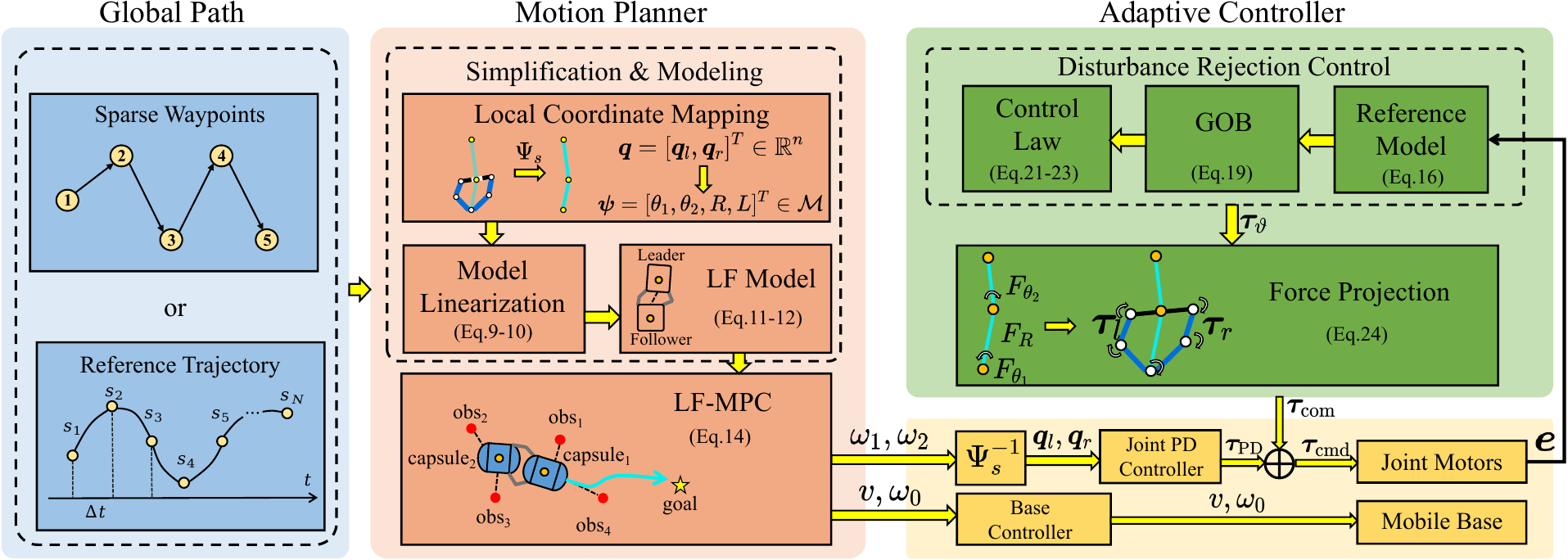}
        \caption{Proposed planning and control framework. The motion planning level generate flexible whole-body motions to track the global path, while the controller level generate control inputs to accurately execute the motions.}
        \label{fig:framework}
\end{figure*}

This section is devoted to the lower-level controller for executing the whole-body motions generated from Section \ref{sec:planning_module}, and mainly focus on the arms’ torques for steering control (as shown in the rightside of Fig. \ref{fig:framework}). During steering, insufficient stiffness in certain directions and disturbances can compromise control robustness, increasing tracking errors and introducing control delays. To achieve better control performance, we first introduce a disturbance rejection control method, which is based on GOB in local coordinates to compensate for disturbances and model uncertainties. Then we propose a force projection method to evenly map the feedforward compensation into the joint space.


\subsection{Disturbance Rejection Control with GOB}

After obtaining the optimal whole-body motions $\boldsymbol{u}*$ with (\ref{optLF}), designing a controller for this nonlinear constrained system is challenging. The primary difficulty lies in steering control. Due to the structure of the dual-arm robot, steering forces must be transmitted through the arm torques, while nonholonomic carts typically have large rotational inertia. This requires higher stiffness in the steering direction compared to other directions. The various disturbances further introduce control uncertainties. Directly applying a joint-level PD controller, as in prior works, often results in insufficient stiffness along the steering direction, whereas simply increasing motor stiffness can induce oscillations. In such cases, using motors with greater torque capacity may seem like the only option, but this increases cost and introduces redundant design complexity. To address these issues, we design an adaptive controller for online compensation of insufficiencies and disturbances.

To adaptively generate compensation force, we employ a model-free disturbance-rejection strategy based on a DOB formulated in the unconstrained local coordinates. The DOB must be capable of estimating disturbances online without relying on accurate system dynamics, as modeling disturbances in cart-pushing tasks is inherently difficult. Moreover, it must capture the nonlinearities associated with the local coordinates. Inspired by \cite{wang2024gob}, we derive a GOB on SE(2) to estimate disturbances. The GOB is a nonlinear variant of the ESO and provides improved accuracy when observing angular variables. In GOB, the sum of model uncertainties and disturbances are modeled as an unified variable  $\boldsymbol{\xi}$, and the second-order dynamics is written as:
\begin{equation}
    \ddot{\boldsymbol{z}}=f(\boldsymbol{z}, \dot{\boldsymbol{z}}_t, \boldsymbol{u}) + \boldsymbol{\xi},
\end{equation}
where $\boldsymbol{z}=[\theta_1, \theta_2, R]^{\top}$. We seperately observe $\hat{\boldsymbol{\xi}} = [\hat{\xi}_{\theta_1}, \hat{\xi}_{\theta_2}, \hat{\xi}_{R}]^{\top}$ and design disturbance rejection control strategy along these directions. Now we introduce the details of the control strategy as below:

\subsubsection{Reference Model} Carts typically have large inertia, so we should avoid accelerating them to speeds that are difficult to control during cart-pushing. Therefore, we introduce interpolation of the desired state in the control strategy.  Let $\boldsymbol{z}_1=\boldsymbol{\bar{z}}_d$ and $\boldsymbol{z}_2=\dot{\boldsymbol{\bar{z}}}_d$ where $\boldsymbol{\bar{z}}_d$ denotes the smoothed desired state, we introduce the Tracking Differentiator (TD) as:
\begin{equation}
\begin{cases}
    \dot{\boldsymbol{z}}_1=\boldsymbol{z}_2 \\
    \dot{\boldsymbol{z}}_2= -\boldsymbol{r}\text{fhan}(\boldsymbol{z}_1-\boldsymbol{z}_d, v_\text{max}) \\ 
\end{cases}
\end{equation}
where $\boldsymbol{r}$ is the gain matrix, $\text{fhan}(\boldsymbol{z}_1-\boldsymbol{z}_d, v_\text{max})$ is defined as in \cite{han2009pid}. TD smooths the desired state, preventing the control instability that would arise from abrupt jumps in $\boldsymbol{z}_d$. The resulting $\boldsymbol{\bar{z}}_d = [\bar{\theta}_{1d}, \bar{\theta}_{2d}, \bar{R}_d]$ is then used in Section \ref{control_law} as the reference input of the feedback control.

\subsubsection{GOB} The linear ESO estimates disturbances through the state estimation error $\boldsymbol{e} = \boldsymbol{z} - \hat{\boldsymbol{z}}$. However, using a simple linear error is inappropriate for angular variables. Therefore, we introduce a geometric error as follows:
\begin{equation}
    \begin{cases}
        e_\theta=\frac{1}{2} [\hat{\boldsymbol{R}}^{\top}\boldsymbol{R}-\boldsymbol{R}^{\top}\hat{\boldsymbol{R}}]^\vee \\
        \dot{\boldsymbol{R}} = \boldsymbol{R}[\omega]_\times
    \end{cases}
\end{equation}
where $\boldsymbol{R}$ is the rotation matrix on SO(2), $[\omega]_\times $ is the skew-symmetric matrix, $[\cdot]^\vee$ is the vee operator and $[x]^\vee_\times=x$. Denoting $\boldsymbol{R}$ as $\boldsymbol{R}(\theta)$, we can further obtain:\begin{equation}
    e_\theta=\frac{1}{2} [\hat{\boldsymbol{R}}^{\top}\boldsymbol{R}-\boldsymbol{R}^{\top}\hat{\boldsymbol{R}}]^\vee=\frac{1}{2}[\boldsymbol{R}(\bar{\theta})-\boldsymbol{R}(-\bar{\theta})]^\vee=\sin \bar{\theta} \label{geoerr}
\end{equation}
where $\bar{\theta} = \theta-\hat{\theta}$. The error calculated using (\ref{geoerr}) is more consistent with the measurement of the angle $\theta_1$ and $\theta_2$.
With the geometric error, we employ a GOB for $\theta_1$ and $\theta_2$ as:
\begin{equation}
    \begin{cases}
\dot{\hat{\boldsymbol{R}}}=\hat{\boldsymbol{R}}[\hat{\omega}+l_1\text{fal}(e_\theta, \sigma_1, \delta_1)]_\times
 \\
\dot{\hat{\omega}}=b_\theta u+l_2\text{fal}(e_\theta,\sigma_2, \delta_2)+\hat{d}_\theta
 \\
 \dot{\hat{d}}_\theta=l_3\text{fal}(e_\theta,\sigma_3, \delta_3)
 \\
\hat{\xi}_\theta=\text{clip}(\hat{d}_\theta, \xi_\text{min}, \xi_\text{max})
\end{cases}
\end{equation}
where $l_1, l_2, l_3$ are the gain parameters, $b_{\theta}$ is the inertia parameter and $\text{fal}$ function is defined as:
\begin{equation}
    \text{fal}(e,\sigma,\delta)=\left\{\begin{array}{ll}
        \frac{e}{\delta^{1-\sigma}}, & \text{if } |e| < \delta \\
        |e|^{\sigma}\text{sign}(e), & \text{otherwise}
    \end{array}\right.
\end{equation}
For radius $R$, we simply apply linear ESO to observe the disturbances. With the GOB, we can obtain the estimated disturbances $\hat{\boldsymbol{\xi}}=[\hat{\xi}_{\theta_1},\hat{\xi}_{\theta_2},\hat{\xi}_{R}]^\top $ as the feedforward term. Now we present the control strategy.
\subsubsection{Control Strategy}\label{control_law} The overall control strategy consists of a feedforward  term and a feedback controller. We employ a nonlinear feedback strategy that provides high stiffness for small errors while deliberately reducing stiffness for large deviations, thereby improving control robustness. To simplify representation, we define $\boldsymbol{\tilde{z}}=[\tilde{\theta}_1, \tilde{\theta}_2, \tilde{R}]^\top$ with $\tilde{\theta} = \sin (\bar{\theta} - \theta)$ and $\tilde{R} = \bar{R} - R$, and denote $\text{fal}(\boldsymbol{\tilde{z}},\boldsymbol{\sigma},\boldsymbol{\delta})$ as applying the $\text{fal}$ operation component-wise to the vector $\boldsymbol{\tilde{z}}$. Then the nonlinear feedback controller can be expressed as:
\begin{equation}
\boldsymbol{u}_z=\beta_1\text{fal}(\boldsymbol{\tilde{z}},\boldsymbol{\sigma}_1,\boldsymbol{\delta}_1) +\beta_2\text{fal}(\dot{\boldsymbol{\tilde{z}}},\boldsymbol{\sigma}_2,\boldsymbol{\delta}_2)   
\end{equation}
where $\boldsymbol{\sigma}_1, \boldsymbol{\sigma}_2,\boldsymbol{\delta}_1, \boldsymbol{\delta}_2 \in \mathbb{R}^3 $ are the vectors of parameters, and
\begin{equation}
    \dot{\boldsymbol{\tilde{z}}}=[\cos(\bar{\theta}_{1d}-\theta_1)\dot{\theta}_1,\cos(\bar{\theta}_{2d}-\theta_2)\dot{\theta}_2,\dot{R}]^{\top}
\end{equation}
Then the total compensation force $\boldsymbol{\tau}_\vartheta$ can be expressed as:
\begin{equation}
    \boldsymbol{\tau}_\vartheta=\boldsymbol{b}_z^{-1}(\boldsymbol{u}_z-\hat{\boldsymbol{\xi}})
\end{equation}
where $\boldsymbol{b}_z=\text{diag}(b_{\theta_1}, b_{\theta_2}, b_R)$ is the gain matrix. The compensation force is generated in the local coordinates based on the observed disturbances $\hat{\boldsymbol{\xi}}$. It effectively mitigates control errors caused by disturbances and insufficient stiffness during steering, while the nonlinearity of the control strategy helps reduce phase lag and oscillations in adaptive control. 
To further eliminate control errors caused by disturbances, we use adaptive control to calculate compensation forces $F_{\theta_1}, F_{\theta_2}, F_R$. To achieve human-like adaptive control without the need of accurate modeling, we employ a model-free disturbance rejection control strategy based on an GOB, that treats complex model uncertainties and external forces as an unified disturbances $\boldsymbol{\xi}$, and obtain the following dynamics:
\vspace{-1mm}
\subsection{Force Projection Method} \label{forcedistribution}

We then project the compensation force into the joint space to obtain the corresponding compensation torques. Utilizing the virtual arm representation in Section \ref{sec:local_coordinate}, we can derive the mapping $h: \mathcal{M} \to \mathbb{R}^{n_q}, \boldsymbol{\tau}_\vartheta \mapsto \boldsymbol{\tau}_{\text{com}}$ as:


\begin{equation}
\boldsymbol{\tau}_{\text{com}}=\begin{bmatrix}\boldsymbol{\tau}_l\\\boldsymbol{\tau}_r\end{bmatrix}=\begin{bmatrix}\eta \boldsymbol{J}_l^{\top}\boldsymbol{\mathcal{J}}_l^{T\dagger} \\ (1-\eta) \boldsymbol{J}_r^{\top}\boldsymbol{\mathcal{J}}_r^{T\dagger}\end{bmatrix}\boldsymbol{\tau}_\vartheta  \label{force_proj}
\end{equation}
where $\eta$ represents the proportion of output between the left and right arms and is set as 0.5, $\boldsymbol{\mathcal{J}}_l, \boldsymbol{\mathcal{J}}_r$ are the Jacobians of the left and right \textit{EE} in virtual arm, $\boldsymbol{J}_l, \boldsymbol{J}_r$ are the Jacobians of the left and right \textit{EE} in the real arms, and $\dagger$ denotes the pseduo inverse (which is equivalent to the regular inverse for 3-DoF manipulators). Equation (\ref{force_proj}) is derived from the principle of virtual work, and the detailed derivation is provided in the \href{https://drive.google.com/file/d/1tkd3y1829YD6_nnqEF2RPGybxyoX_pTm/view}{supplementary material}. In practice, the compensation torque $\boldsymbol{\tau}_{\text{com}}$ is combined with a low-gain joint PD controller, and the overall control torque $\boldsymbol{\tau}_\text{cmd}$ of the dual-arm is $\boldsymbol{\tau}_\text{cmd} = \boldsymbol{\tau}_\text{PD}+\boldsymbol{\tau}_\text{com}$.

\section{Experiment}
\label{exp_sec}

In this section, simulations and real-world experiments validate our motion planning method and adaptive controller for cart-pushing. First, we compare our motion planner to baseline algorithms on steep reference trajectories tracking tasks in simulation. Then, we design motions execution tasks under different payloads to evaluate the disturbance rejection capability and accuracy of the controllers. Finally, we conduct navigation experiments in challenging scenarios to assess the maneuverability of the proposed framework and its control robustness under extreme conditions.

\begin{figure}[!t]
\centerline{\includegraphics[width=\columnwidth]{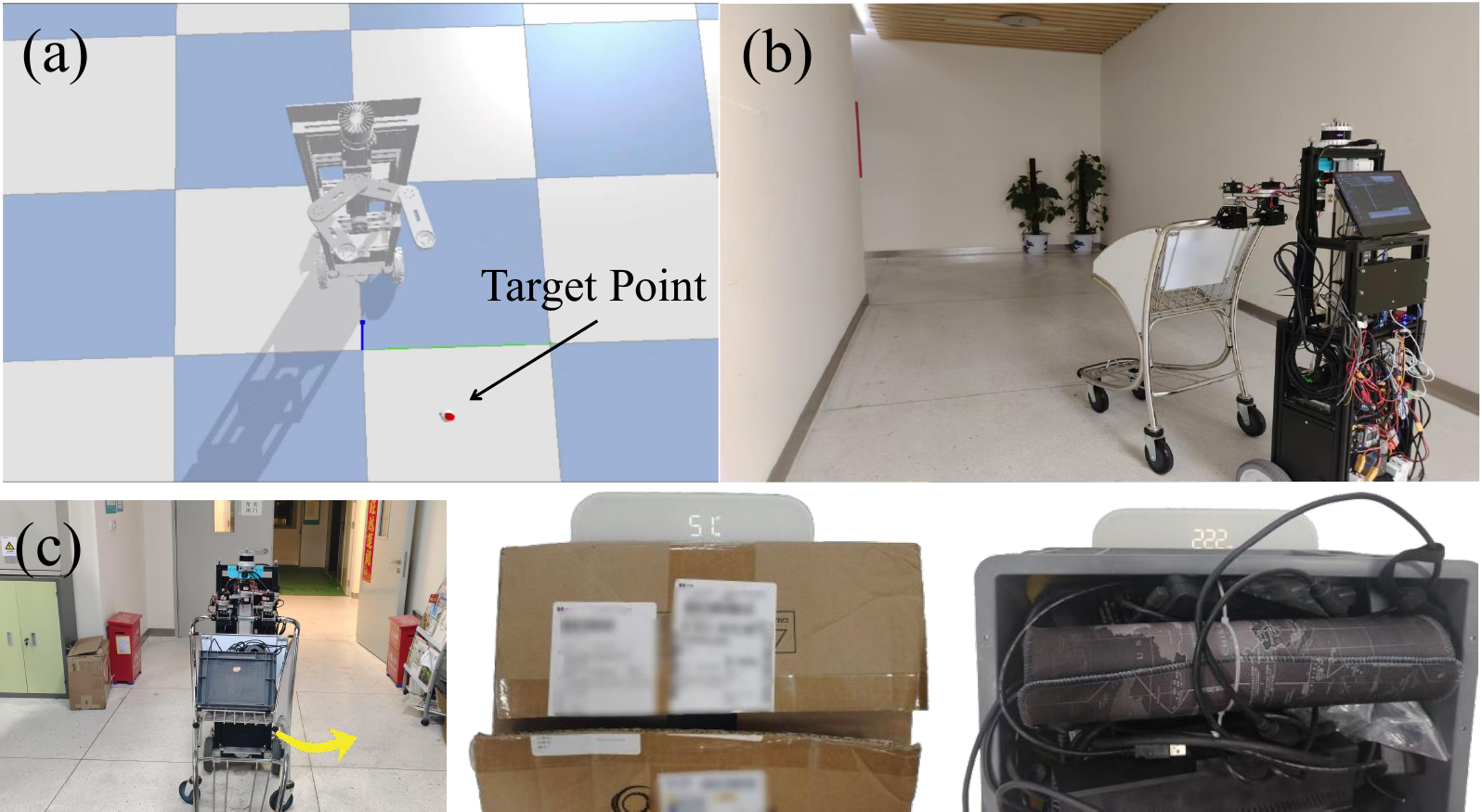}}
\caption{Experiments Setup. (a) Simulation. (b) Navigation in narrow passage. (c) Adaptive control with payloads. The payloads in experiments are 5.1 kg and 22.2 kg.}
\label{exp_overview}\vspace{-1mm}
\end{figure}

\begin{figure*}
     \centering
     \includegraphics[width=0.9\textwidth]{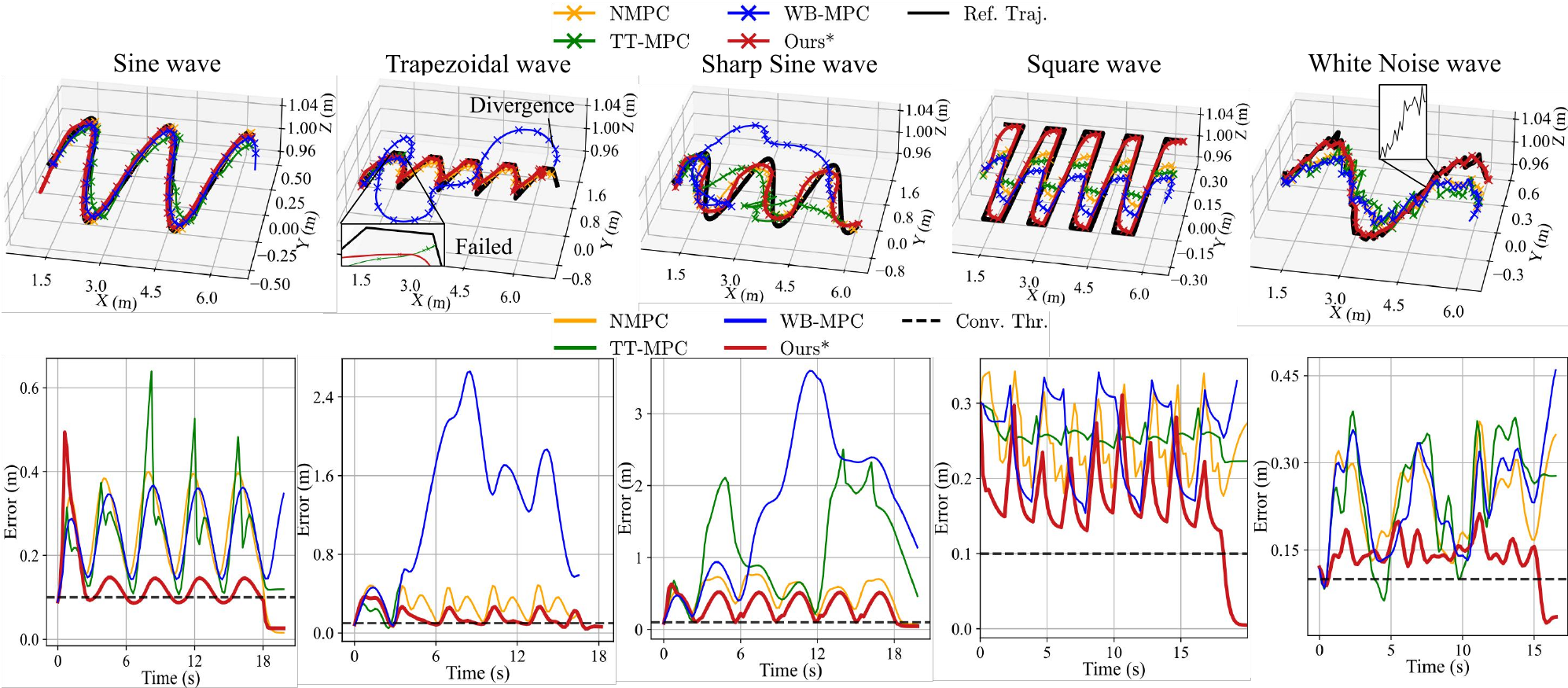}

     \caption{Visualization of the trajectories and tracking errors of 5 different reference trajectories tracking tasks.}
     
    \label{fig:mpresult}
\end{figure*}

\subsection{Experiment Setup}

For simulations, the computation and visualization of the algorithms are carried out on a laptop computer with Intel Core i7-11657 CPU@4.70GHz, and is implemented in Python in Ubuntu 20.04 and ROS Noetic. For real-world experiments, the algorithm is implemented in C++11 in Ubuntu 20.04 and ROS2 Galactic. All planners in Section \ref{exp:motion_planning} are formulated in CasADi \cite{andersson2019casadi} and solved with the IPOPT solver. The overall framework runs at a frequency of 22 Hz.

\subsection{Motion Planning Experiments}\label{exp:motion_planning}

For the motion planning experiments, we design challenging reference trajectories and static pose tracking tasks to evaluate the flexibility of different algorithms during whole-body coordination. We evaluate planners with different reduced-order models and compare their real-time tracking errors and convergence accuracy under various push-pose strategies. These strategies differ in how effectively they utilize the redundant DoFs. For comparison, NMPC is an online optimization method that keeps the arm fixed at Centered Pose and controls only the base’s forward and steering motions, as shown in \cite{xiao2022robotic}. WB-MPC plans whole-body motion directly on full-order model, which is similar to \cite{schulze2023trajectory}. TT-MPC and LF-MPC are the proposed optimization methods based on the TT and LF models, respectively, as defined in (\ref{ttmodel}) and (\ref{LF-model}). All the planners shares the same parameters as $\boldsymbol{Q} = \mathrm{blkdiag}\big(\mathrm{diag}(2,2,0.5),\,0.1\,\boldsymbol{I}_{n_x-3}\big)$, $\boldsymbol{R} = 0.1\,\boldsymbol{I}_{n_u}$ and $N=50$ where $n_x, n_u$ are the dimension of the state and controls. In our experiments, based on physical measurements, $L_1=L_{\text{cap}}=0.35\text{m}, L_2 = r_{\text{cap}}=0.2\text{m}$. The experiments are conducted solely in simulation, as it allows easier measurement of position errors compared to real-world tests, which is sufficient for validating the performance of the motion planning. We will demonstrate that insufficient whole-body coordination can lead to large tracking errors or even divergence in local planning for cart-pushing.


We conduct reference trajectory tracking tasks and record the time varying tracking error with different planners. The chosen reference trajectories are sharp and nonsmooth, and even with strong noise. We calculate the tracking error by $e=2\left \| \boldsymbol{p}_d-\boldsymbol{p}_c \right \|^2+0.5\left \| \theta_d-\theta_c \right \|^2$, and the results are shown in Fig. \ref{fig:mpresult}. Due to the large turning radius, the other methods exhibit cusp points (local reversals) or even divergence when tracking steep trajectories. NMPC shows relatively large tracking errors across all tasks. WB-MPC diverges in the Trapezoidal Wave and Sharp Sine Wave tasks. TT-MPC reduces the tracking error but fails in the Trapezoidal Wave task due to planning timeouts. In contrast, LF-MPC leverages whole-body coordination to track steep trajectories smoothly with small tracking errors, demonstrating superior flexibility. LF-MPC can run at an average frequency of 80 Hz, ranging from 50 to 100 Hz in these experiments. Results in Section \ref{exp:nav} further demonstrates that LF-MPC operates at approximately 22 Hz with obstacle-avoidance constraints, satisfying real-time requirements.

\begin{figure}
     \centering
     \includegraphics[width=0.95\columnwidth]{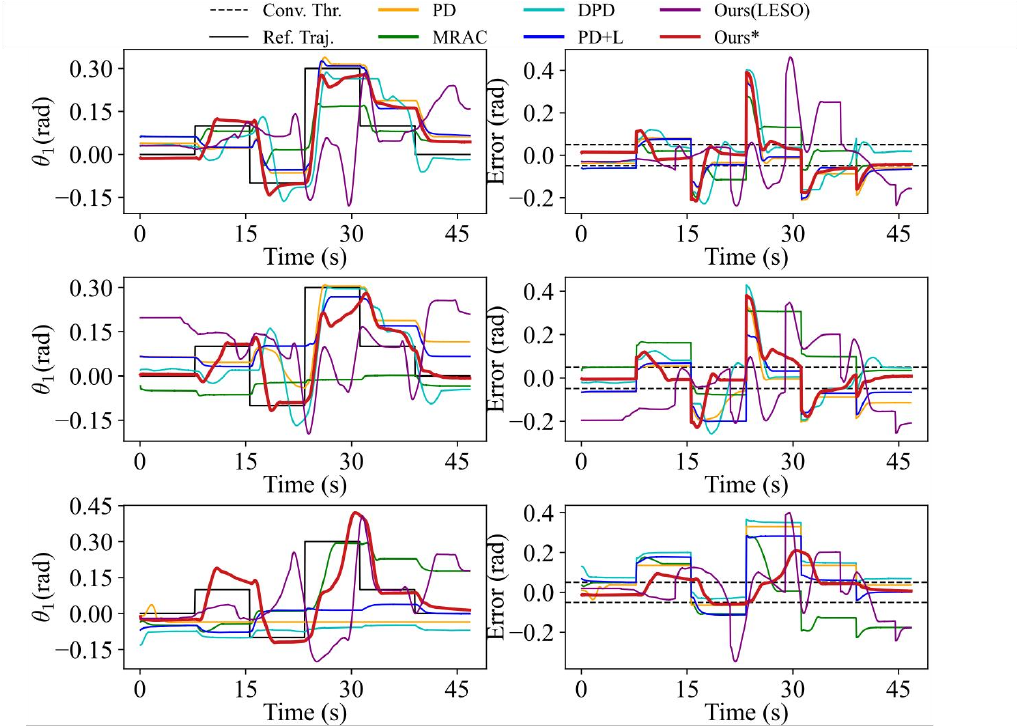}
     \caption{Experimental results of controllers tracking performance on reference $\theta_1$ trajectory with different payloads. (a) Without payload. (b) With 5.1kg payload. (c) With 22.2kg payload.}
    \label{exp:force}\vspace{-1mm}
\end{figure}

\subsection{Adaptive Control Experiments}\label{exp:adaptive-control}

To evaluate the robustness of our Adaptive Controller under disturbances, we design a control task in which the controller execute a desired trajectory, and compare the execution error ($\mathrm{rad}$) against several baseline controllers. We design a trajectory for $\theta_1$ as the desired reference, where a steep change is introduced at a specific time. The controllers becomes more prone to lag or overshoot under heavier payloads. This tests the controllers’ convergence speed and stability. The baselines we select are advanced controllers that do not rely heavily on accurate models, reflecting the practical difficulty of obtaining precise perception and modeling of carts under varying payload conditions. The baseline controllers are namely:



\begin{itemize}
    \item \textbf{PD}: A configuration space PD controller, namely $\boldsymbol{\tau}_{\text{PD}}=\boldsymbol{K}_p\Tilde{\boldsymbol{q}}+\dot{\boldsymbol{K}_d\Tilde{\boldsymbol{q}}}$. We set $\boldsymbol{K}_p=15$, which is the maximum allowable stiffness before motor oscillation occurs, and $\boldsymbol{K}_d=\sqrt{30}$ according to  $\boldsymbol{K}_d=\sqrt{2\boldsymbol{M}\boldsymbol{K}_p}$ where $\boldsymbol{M}=1.0$ is the motor inertia.
    \item \textbf{PD+L}: A Joint Space PD controller compensated by a \textbf{L}ocal coordinate PD controller $\boldsymbol{\tau}_{\text{com}}=\boldsymbol{K}_\mathcal{P} (\Tilde{\boldsymbol{z}})-\boldsymbol{K}_\mathcal{D}\dot{\Tilde{\boldsymbol{z}}}$ as compensation, where $\Tilde{\boldsymbol{z}}=[\Tilde{\theta}_1,\Tilde{\theta}_2,\Tilde{R}]^T$ and $\Tilde{\theta}_1,\Tilde{\theta}_2$ can be calculated with (\ref{geoerr}). The velocity $\dot{\boldsymbol{z}}=[\dot{\theta}_1, \dot{\theta}_2, \dot{R}]^T$ is estimated by an EKF. The command torque is calculated as
    $\boldsymbol{\tau}_\text{cmd}=\boldsymbol{\tau}_\text{PD}+\boldsymbol{\tau}_\text{com}$. We set a relatively small stiffness in the configuration space ($\boldsymbol{K}_p=5$) and a larger stiffness in local coordinate ($\boldsymbol{K}_\mathcal{P}=20$), and $\boldsymbol{K}_\mathcal{D}=2\sqrt{10}$ is tuned according to the control performance.
    \item \textbf{DPD}: A dual–loop PD controller formulated in local coordinates \cite{roy2002adaptive}, commonly used in legged–robot body stabilization. 
The outer loop regulates the desired local position based on the position error, while the inner loop tracks the reference $\boldsymbol{z}_d$ through PD feedback. 
The resulting desired joint configuration $\boldsymbol{q}_d$ is obtained via the mapping in (\ref{force_proj}), and the joint torques are generated through a joint–space PD controller $\dot{\boldsymbol{z}}_d = \boldsymbol{K}_{\mathcal I} \left( \boldsymbol{z} - \boldsymbol{z}_d \right)$, $\ddot{\boldsymbol{z}}_d = 
        \boldsymbol{K}_{\mathcal P}\left(\boldsymbol{z}_d - \boldsymbol{z}\right)
        + \boldsymbol{K}_{\mathcal D}\left(\dot{\boldsymbol{z}}_d - \dot{\boldsymbol{z}}\right)$ and $\boldsymbol{q}_d = \Psi_s^{-1}(\boldsymbol{\psi})$.

\item \textbf{MRAC:} A model-reference adaptive controller \cite{tung2000application}. 
The reference model is $\dot{\mathbf z}_m = \mathbf A_m \mathbf z_m + \mathbf B_m \mathbf z_d$, in which:
\begin{equation}
    \mathbf A_m =
    \begin{bmatrix}
        0 & 1 \\
        -a_m^2 & -2a_m
    \end{bmatrix}, \quad
    \mathbf B_m =
    \begin{bmatrix}
        0 \\
        b_m
    \end{bmatrix}.
\end{equation}
The control law is parameterized as $\boldsymbol{\tau}_\vartheta = \mathbf K_z \mathbf z + \mathbf K_r \mathbf z_m$,
and the adaptive gains $\mathbf K_z$ and $\mathbf K_r$ are updated online using a Lyapunov-based rule. 
A first-order low-pass filter is applied to the reference $\mathbf z_d$.
\end{itemize}

To evaluate robustness under various disturbance, all controller parameters were first optimally tuned in the no-payload condition and then kept fixed as additional loads were introduced. For GOB, gain-related parameters such as $l_1, l_2, l_3$ are tuned to be as large as possible before noise amplification appears in the disturbance estimation, while other range-related parameters are adjusted according to experimental requirements. We set $l_1=12, l_2=48, l_3=60$, $\sigma_1 = \sigma_2 =0.2$, $\delta_1 = \delta_2 = 0.5$, $\sigma_3 = 0.5$, $\delta_3 = 0.7$, $\xi_\text{min} = -3.0$ and $\xi_\text{max} = 3.0$ in our experiment. As shown in Fig.~\ref{exp:force}, most algorithms perform well without payloads. However, once a load is applied, the tracking performance of the PD controller degrades significantly. The PD+L controller shows marginal improvement but still fails to keep the error within an acceptable bound. The remaining controllers satisfy the convergence requirements, though the DPD controller exhibits noticeable lag. Under heavy payloads, nearly all methods fail except ours, which maintains high tracking accuracy with minimal lag. In practice, the GOB and our adaptive controller runs at over 3000 Hz. We also compare the tracking errors of our adaptive controller using the LESO (purple) and the GOB (red). The GOB exhibits smaller oscillations and faster convergence, which can be attributed to its more accurate disturbance estimation when the angular error is large.

We further evaluate the regulation performance of different controllers under varying payloads. The desired trajectory is defined over $\theta=\pm 0.3 \mathrm{rad}$. A 22.2kg load is applied at specific time instants, and once the controllers’ adaptive parameters have converged, the load is removed. Using PD and DPD as baselines, we demonstrate that the proposed controller achieves rapid and stable regulation. The demonstration video is available on our 
\href{https://sites.google.com/view/mpac-pushing/}{website}.
\begin{figure*}
     \centering
     \includegraphics[width=0.9\textwidth]{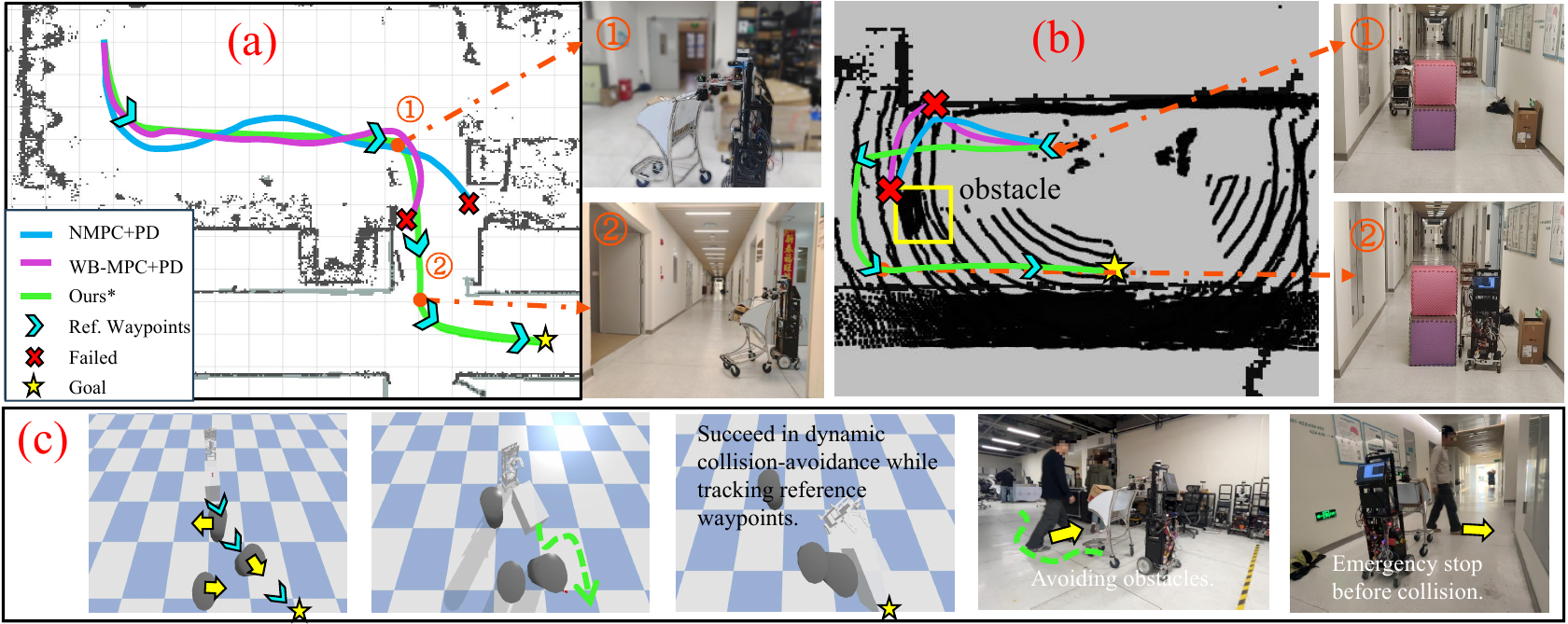}
     \caption{Result of navigation task. (a) Long-Distance Navigation. (b) Narrow-Space Navigation. (c) Dynamic Obstacle Avoidance Task in both simulation and real-world setup.
}
    \label{exp:track}
\end{figure*}
\subsection{Robot Navigation Task} \label{exp:nav}
We conduct 3 challenging cart-pushing tasks to validate the performance of the proposed framework, namely long-distance navigation, narrow-space navigation, and dynamic obstacle avoidance tasks, to show the framework’s overall maneuverability. These tasks require not only the flexibility of the planner but also the precision of the controller’s execution. The long-distance navigation task requires the cart to traverse multiple sharp turns and exit the laboratory through a narrow door. The narrow-space navigation task involves maneuvering around obstacles and performing U-turns within limited space, which imposes high demands on system maneuverability. The dynamic obstacle-avoidance task requires pushing the cart through a crowded environment toward a goal pose while avoiding collisions with pedestrians.

\begin{figure}
     \centering
     \includegraphics[width=\columnwidth]{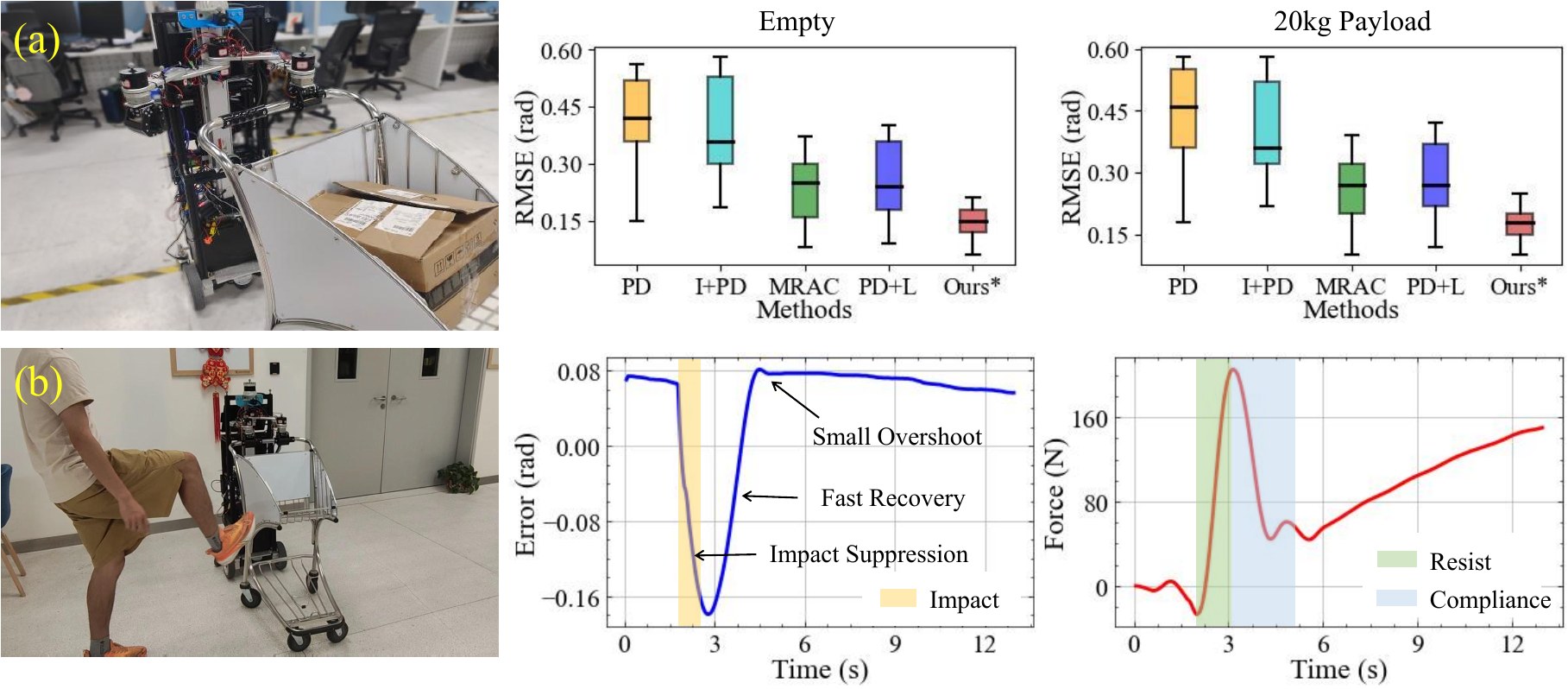}
     \caption{Extreme conditions tests. (a) Transportation task with one arm failure. (b) Impact resistance test.}
    \label{exp_extreme}
    \vspace{-2mm}
\end{figure}
For the experiments setup, we employ the FAST-LIO2 \cite{xu2022fast} algorithm for localization. We further employ the Generalized Iterative Closest Point (GICP) \cite{segal2009generalized} algorithm to match the received LiDAR point cloud and the established 3D map point cloud for initial pose calculation. We use the Octomap \cite{hornung2013octomap} algorithm to diminish the height of a 3D map to 2D cost map, and the corresponding cost map is yielded by expanding the obstacle pixels. We adopt the Hybrid A* algorithm with the TT model as the global planner to generate collision-free paths, which are then sparsely sampled to form the reference trajectories (detailed deployment is provided in the \href{https://drive.google.com/file/d/1tkd3y1829YD6_nnqEF2RPGybxyoX_pTm/view}{supplementary material}). We implement three planning and control frameworks for comparison: NMPC+PD, WB-MPC+PD (corresponding to~\cite{xiao2022robotic} and~\cite{schulze2023trajectory}), and our proposed framework for local planning and control. The parameters of the planners and controllers are set as in previous experiments, and we incorporate CBF-based constraints in all the planners for collision avoidance. NMPC+PD and WB-MPC+PD solve the local planning optimization using an oversimplified model and a full-order model, respectively, and employ standard non-adaptive controllers (\textbf{PD} in Section \ref{exp:adaptive-control}) for dual-arm control. In contrast, our method leverages the proposed reduced-order model (LF model) and incorporates adaptive control to compensate for disturbances. In the long-distance navigation task, baseline methods (blue and purple) fail to execute the sharp turns and consequently collide with the environment. In contrast, our method (cyan) successfully performs 3 sharp turns and reaches the goal pose. In the narrow-space navigation task, only our method achieves whole-body coordination to sufficiently reduce the turning radius, enabling a full turnaround within the confined corridor.
In the dynamic obstacle-avoidance task, obstacle point clouds are obtained from the LiDAR sensor and filtered based on their distance to the base link and height, yielding a clean obstacle point cloud. We then further select the 5 closest points, measured using the SDF distance defined in (\ref{sdf_capsule}), to construct the CBF constraints. In the experiments, $d_{\text{safe}}$ is set to 0.1m. The robot is commanded to track a long reference trajectory in a complex narrow space, while multiple pedestrians cross the trajectory to create dynamic obstructions. As shown in the demonstration video on the \href{https://sites.google.com/view/mpac-pushing/}{website}, the robot exhibits reactive collision avoidance behavior while tracking the reference trajectory in both simulation and real-world experiments.

\subsection{Robustness Testing under Extreme Conditions}\vspace{-1.5mm}
To demonstrate the robustness of our algorithm, we tested two extreme scenarios. In the first, we simulated a single-arm failure, requiring the robot to complete the transportation task using only the remaining arm. Using the framework described in Section \ref{exp:nav}, the motion planner generated trajectories for the robot’s arms and base based on 10 reference waypoints. We evaluated different controllers controlling the single arm under both no-load and 22.2 kg payload conditions, measuring accuracy and response delay via Root Mean Squared Error (RMSE). Results show that our method achieves superior control accuracy and significantly outperforms other controllers in terms of stability.

The second scenario involved an impact resistance test. Impacts typically cause significant disturbances to the controller \cite{chen2024compliance}. Insufficient arm stiffness during impact can lead to uncontrollable acceleration, while overly aggressive disturbance compensation may cause severe overshoot. In our test, a sudden lateral impact force of approximately 200N was applied to the cart, causing a sharp but brief disturbance to the controller. As shown in Fig. 11, we recorded the trajectory of $\theta_1$ and the corresponding error. Thanks to the nonlinear control law, our controller switches to a resist mode upon impact, rapidly increasing stiffness to stabilize the system and quickly returning to the desired state. After the error decreases, the system transitions to a compliance mode with minimal overshoot due to constraints imposed by the reference model.

\section{Conclusion} \label{conclusion}

In this work, we investigate efficient whole-body coordination to enhance maneuverability in cart pushing tasks, and robust adaptive control that handles complex dynamics, variable payloads, and external disturbances for more challenging scenarios. We systematically analyze task requirements for better control performance, designing a motion planning algorithm that generates flexible motions. An adaptive control strategy is proposed to increase control accuracy and cope with complex disturbances. Adequate simulations and experiments validate the flexibility and stability of our method.

Note that the proposed algorithm makes no assumptions about the cart configuration or the manipulator's DoFs. Theoretically, it can be extended to various cart-like objects (e.g.,  platform carts and wheelchairs) and hardware platforms.  Future work will focus on extending the algorithm to a wider range of carts and robots for further validation.

\bibliographystyle{IEEEtran}

\bibliography{reference}

\vspace{-15mm}
\begin{IEEEbiography}
[{\includegraphics[width=1in,height=1.25in,clip,keepaspectratio]{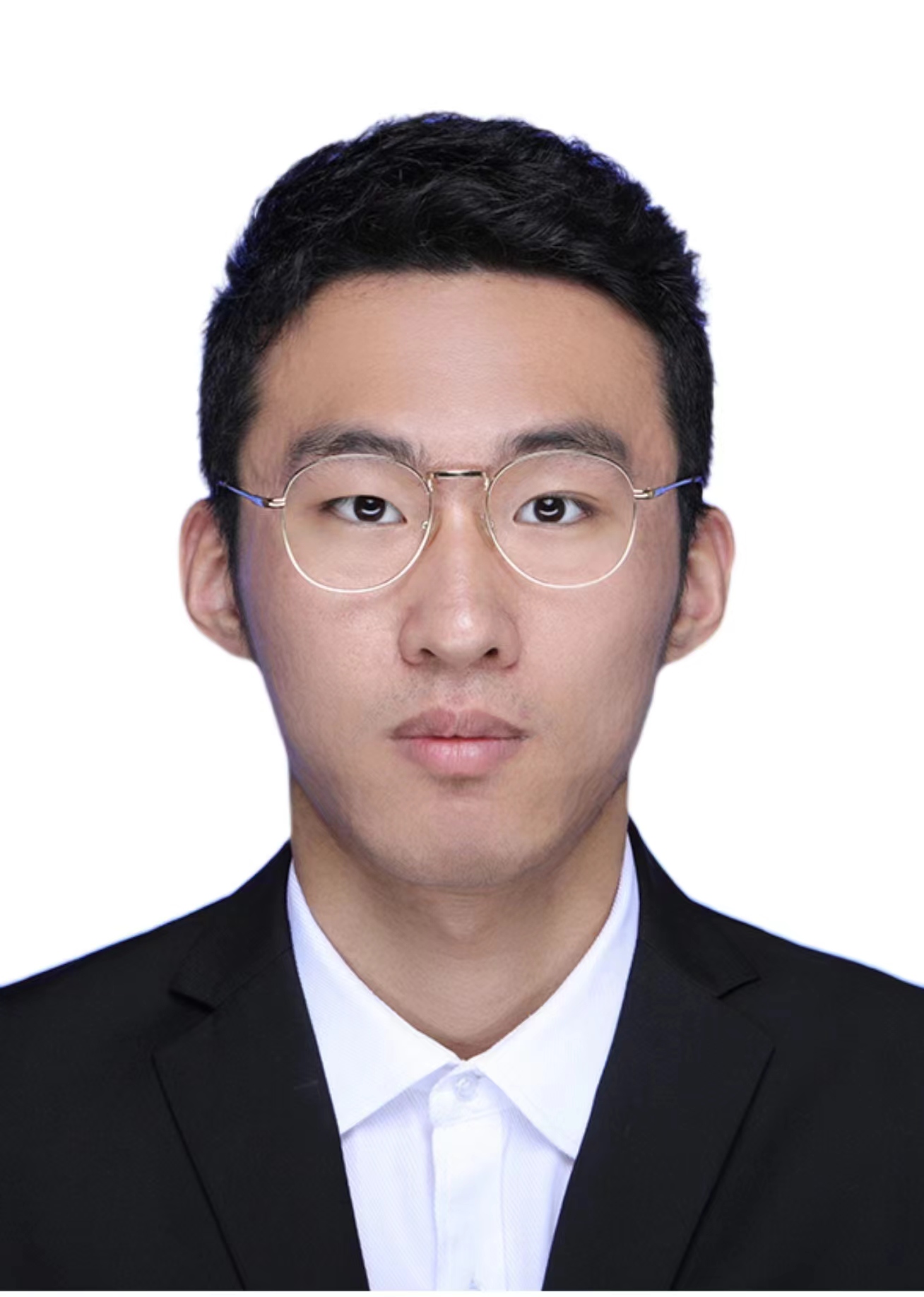}}]
{Zhe Zhang} received his B.E. degree in Electrical Engineering from Beijing Jiaotong University, Beijing, China, in 2022. He is currently pursuing the Ph.D. degree with the Department of Electronic and Electrical Engineering, Southern University of Science and Technology, Shenzhen, China. His research interests include motion planning, optimal control and human-robot interaction.
\vspace{-10mm}
\end{IEEEbiography}

\begin{IEEEbiography}
[{\includegraphics[width=1in,height=1.25in,clip,keepaspectratio]{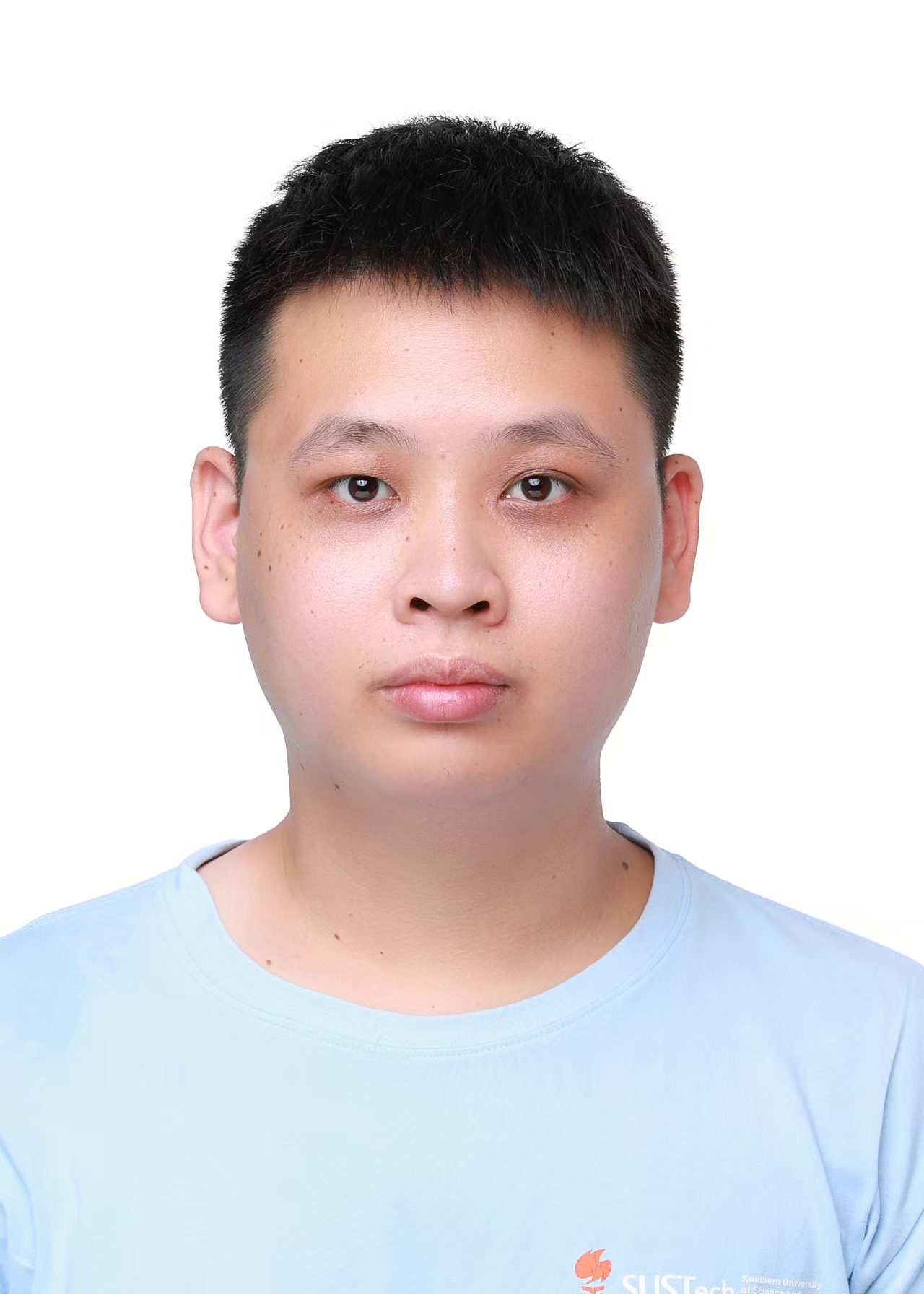}}]
{Peijia Xie} received the B.E. degree in Electronic Information Engineering from the School of Physics and Telecommunications Engineering, South China Normal University, Guangzhou, China, in 2022. He is currently pursuing the Master degree with the Department of Electronic and Electrical Engineering, Southern University of Science and Technology, Shenzhen, China. His research interests autonomous driving.
\vspace{-10mm}
\end{IEEEbiography}

\begin{IEEEbiography}
[{\includegraphics[width=1in,height=1.25in,clip,keepaspectratio]{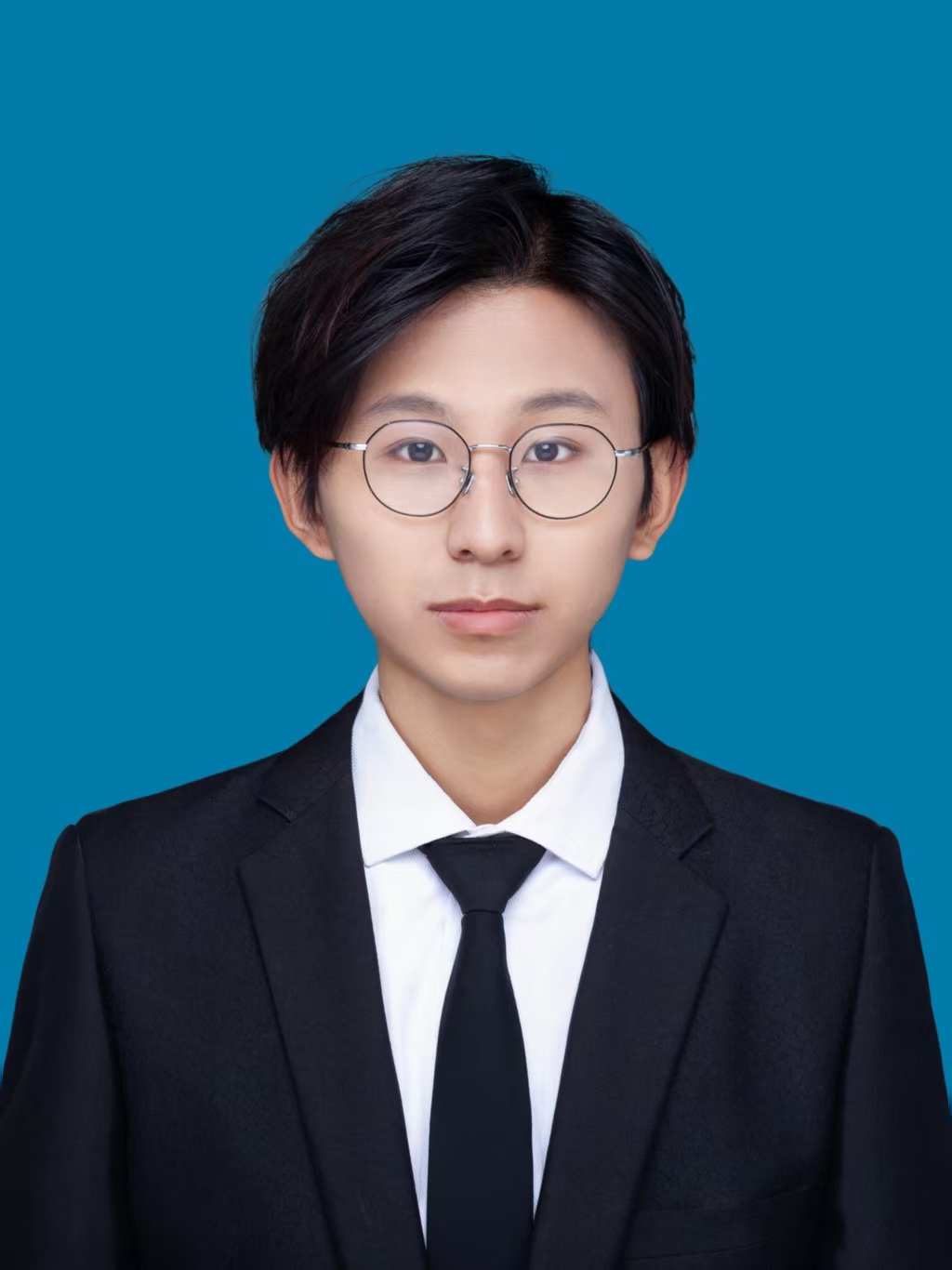}}]
{Yuhan Pang} received the B.E. degree in Measurement and Control Technology and Instrument from the School of Mechanical and Electronic Engineering, Wuhan University of Technology, Wuhan, China, in 2024. He is currently pursuing the Master degree with the Department of Electronic and Electrical Engineering, Southern University of Science and Technology, Shenzhen, China. His research interests mobile robot navigation.
\vspace{-10mm}
\end{IEEEbiography}
\begin{IEEEbiography}
[{\includegraphics[width=1in,height=1.25in,clip,keepaspectratio]{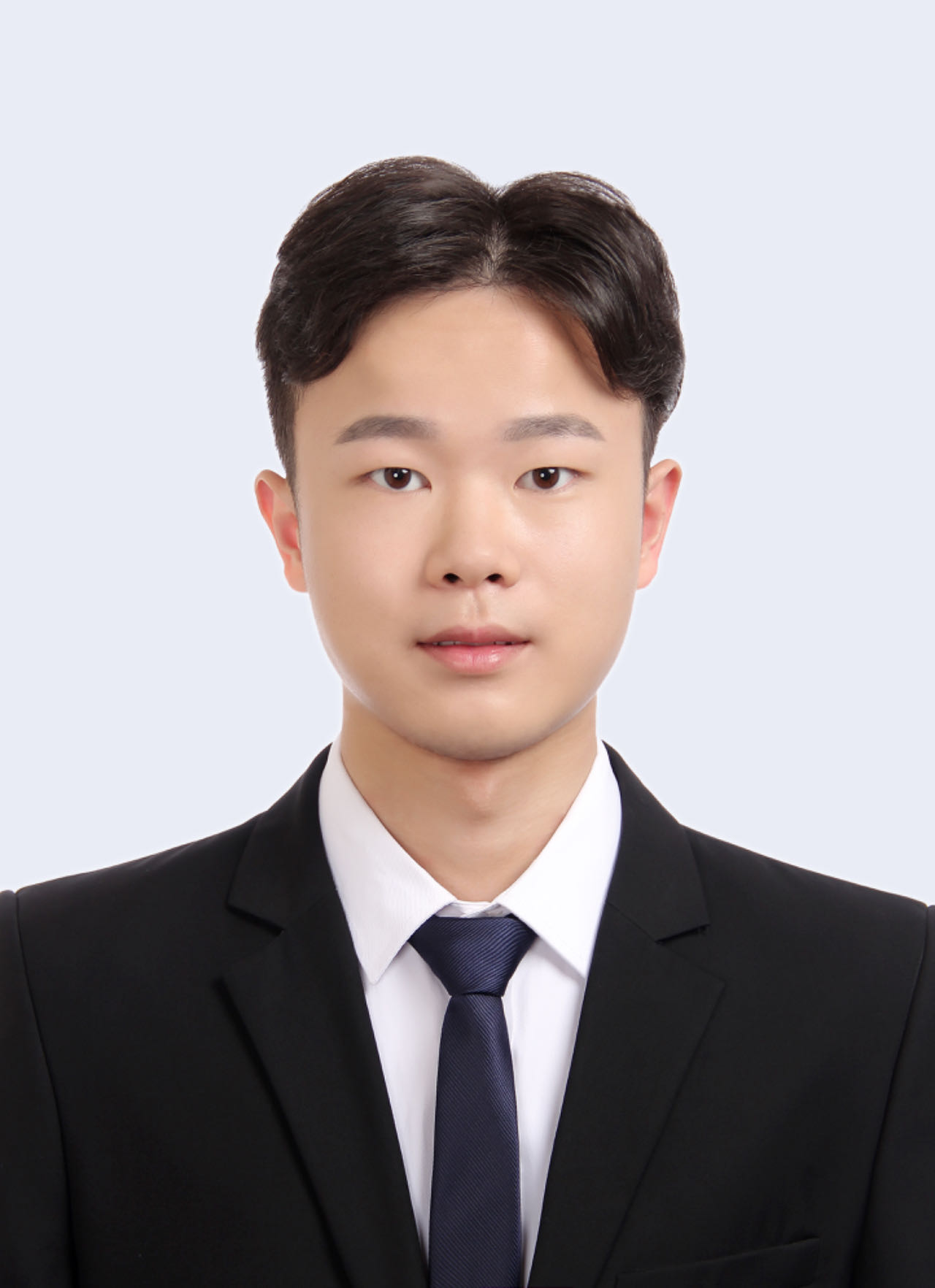}}]
{Zhirui Sun} received the B.E. degree in information engineering from the Department of Electronic and Electrical Engineering, Southern University of Science and Technology, Shenzhen, China, in 2019. He is currently pursuing the Ph.D. degree with the Department of Electronic and Electrical Engineering, Southern University of Science and Technology, Shenzhen, China. His research interests include robot perception and motion planning.
\vspace{-10mm}
\end{IEEEbiography}

\begin{IEEEbiography}
[{\includegraphics[width=1in,height=1.25in,clip,keepaspectratio]{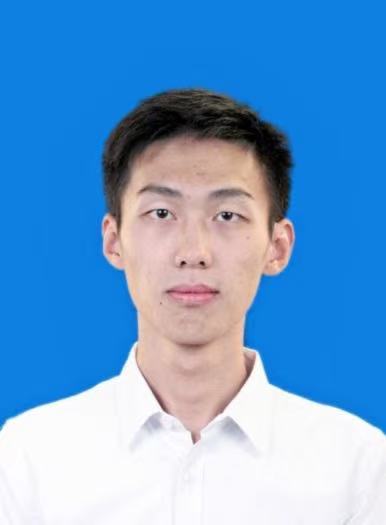}}]
{Bingyi Xia} received the B.E. degree in microelectronics science and engineering from the Southern University of Science and Technology, Shenzhen, China, in 2020. He is currently pursuing the Ph.D. degree with the Department of Electronic and Electrical Engineering, Southern University of Science and Technology, Shenzhen, China. His research interests include robot motion planning.
\vspace{-10mm}
\end{IEEEbiography}

\begin{IEEEbiography}
[{\includegraphics[width=1in,height=1.25in,clip,keepaspectratio]{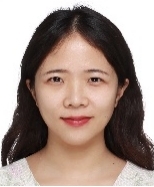}}]
{Bi-Ke Zhu} received the B.E. degree in mechanical engineering from Dalian University of Technology, Dalian, China, in 2019, and Ph.D. degree in mechanical engineering from Shanghai Jiao Tong University, Shanghai, China, in 2024. She is currently a Postdoc researcher with the Department of Electronic and Electrical Engineering, Southern University of Science and Technology, Shenzhen, China. Her research interests include motion planning and intelligent control of robotic systems.
\vspace{-10mm}
\end{IEEEbiography}

\begin{IEEEbiography}
[{\includegraphics[width=1in,height=1.25in,clip,keepaspectratio]{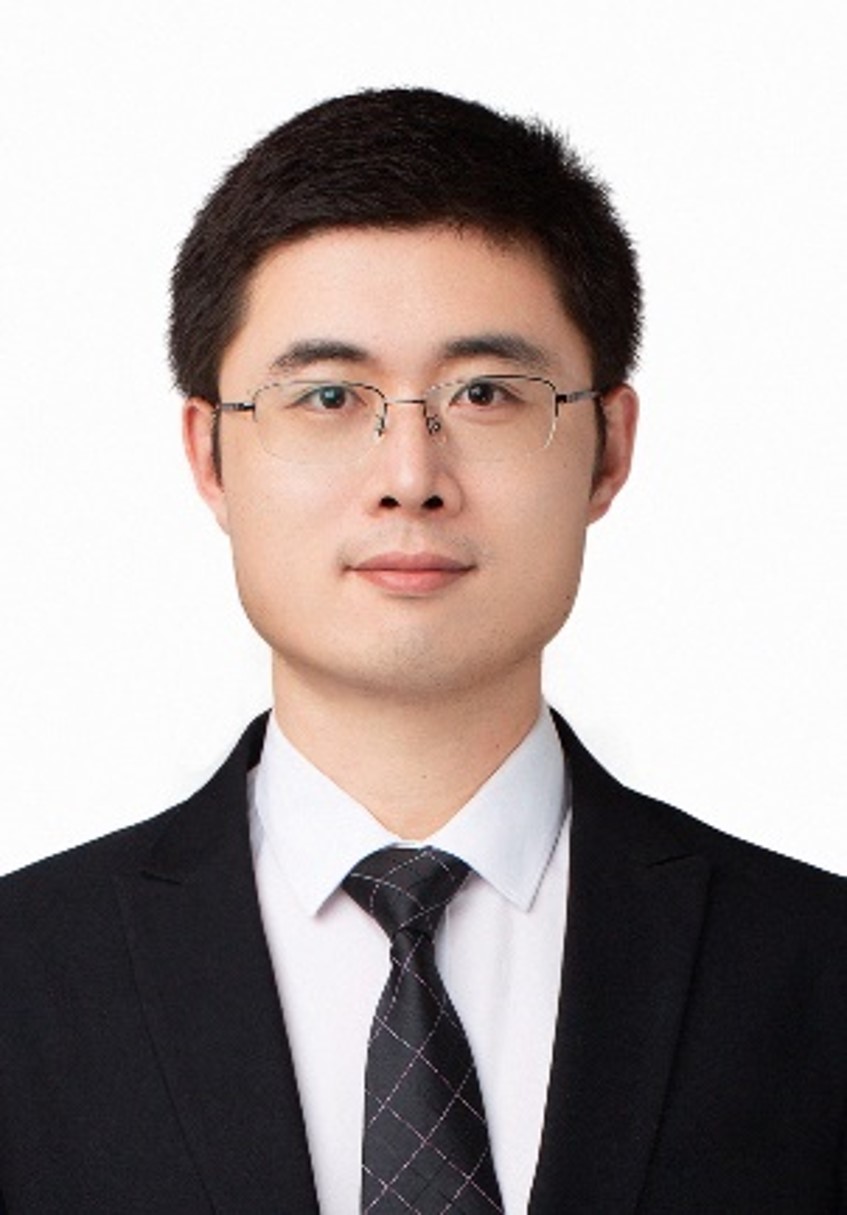}}]
{Jiankun Wang} (Senior Member, IEEE) received the B.E. degree in automation from Shandong University, Jinan, China, in 2015, and the Ph.D. degree from the Department of Electronic Engineering, The Chinese University of Hong Kong, Hong Kong, in 2019.

He is currently an Assistant Professor with the Department of Electronic and Electrical Engineering, Southern University of Science and Technology, Shenzhen, China. His current research interests include motion planning and control, human–robot interaction, and machine learning in robotics.

Currently, he serves as the associate editor of IEEE Transactions on Automation Science and Engineering, IEEE Transactions on Intelligent Vehicles, IEEE Robotics and Automation Letters, International Journal of Robotics and Automation, and Biomimetic Intelligence and Robotics.
\end{IEEEbiography}

\end{document}